\documentclass[a4paper,fleqn]{cas-sc}
\usepackage{dblfnote}
\usepackage[]{natbib}
\usepackage{xcolor}
\usepackage{hyperref}
\usepackage{algorithm}
\usepackage{algorithmic}
\usepackage{multicol}

\usepackage{subfig,graphicx}
\usepackage[T1]{fontenc}
\usepackage{mathdesign}
\usepackage{lipsum}
\usepackage{titlesec}
\newcommand{\charternumbers}{\fontfamily{bch}\selectfont}
\DeclareTextFontCommand{\textcharter}{\charternumbers}

\usepackage{tabularx}

\titleformat{\section}{\normalfont\fontsize{9pt}{12pt}\bfseries}{\textcharter{\thesection.}}{0.5em}{}
\titleformat{\subsection}{\normalfont\fontsize{8pt}{12pt}\itshape}{\textcharter{\thesubsection.}}{0.3em}{}
\titleformat{\subsubsection}{\normalfont\fontsize{8.3pt}{12pt}\itshape}{\textcharter{\thesubsubsection.}}{0.4em}{}

\titleformat{name=\section,numberless}{\normalfont\fontsize{9pt}{12pt}\bfseries}{}{0em}{}
\titleformat{name=\subsection,numberless}{\normalfont\fontsize{9pt}{12pt}\bfseries}{}{0em}{}
\titleformat{name=\subsubsection,numberless}{\normalfont\fontsize{9pt}{12pt}\bfseries}{}{0em}{}

\usepackage{hyperref}

\usepackage{balance}
\usepackage{color}
\hypersetup{
  colorlinks=true,
  linkcolor=cyan,
  urlcolor=cyan,
  citecolor=cyan,
  linktoc=all
}
\usepackage{fancyhdr}
\def\tsc#1{\csdef{#1}{\textsc{\lowercase{#1}}\xspace}}
\tsc{WGM}
\tsc{QE}
\tsc{EP}
\tsc{PMS}
\tsc{BEC}
\tsc{DE}

\begin{document}
\thispagestyle{empty}

\let\WriteBookmarks\relax
\def\floatpagepagefraction{1}
\def\textpagefraction{.001}
\let\printorcid\relax

\shorttitle{NGI-Attack}


\title[mode = title]{Improving Adversarial Transferability with Neighbourhood Gradient Information}  



\author[1]{\textcolor{black}{Haijing Guo}}
\fnmark[1] 
\ead{hjguo22@m.fudan.edu.cn}

\author[1]{\textcolor{black}{Jiafeng Wang}}[type=editor,
    auid=000,bioid=2,
    role=,]
\fnmark[1] 
\ead{jiafengwang21@m.fudan.edu.cn} 

\author[2,3]{\textcolor{black}{Zhaoyu Chen}}[style=chinese]
\fnmark[2] 
\cormark[1]
\ead{zhaoyuchen20@fudan.edu.cn}

\author[2,3]{\textcolor{black}{Kaixun Jiang}}
\ead{kxjiang22@m.fudan.edu.cn}

\author[1]{\textcolor{black}{Lingyi Hong}}
\ead{lyhong22@m.fudan.edu.cn}

\author[2,3]{\textcolor{black}{Pinxue Guo}}
\ead{pxguo21@m.fudan.edu.cn}

\author[2,3]{\textcolor{black}{Jinglun Li}}
\ead{jingli960423@gmail.com}

\author[1,2,3]{\textcolor{black}{Wenqiang Zhang}}
\cormark[1]
\ead{wqzhang@fudan.edu.cn}

\address[1]{Shanghai Key Lab of Intelligent Information Processing, School of Computer Science, Fudan University, Shanghai 200433, China}
\address[2]{Shanghai Engineering Research Center of AI \& Robotics, Academy for Engineering \& Technology, Fudan University, Shanghai 200433, China}
\address[3]{Engineering Research Center of Robotics, Ministry of Education, Academy for Engineering \& Technology, Fudan University, Shanghai 200433, China}

\fntext[1]{Equal contribution.} 
\cortext[1]{Corresponding author.} 
\fntext[2]{Project lead.}

\begin{abstract}
Deep neural networks (DNNs) are known to be susceptible to adversarial examples, leading to significant performance degradation. In black-box attack scenarios, a considerable attack performance gap between the surrogate model and the target model persists. This work focuses on enhancing the transferability of adversarial examples to narrow this performance gap. 
We observe that the gradient information around the clean image, \textit{i.e., Neighbourhood Gradient Information (NGI)}, can offer high transferability. 
Based on this insight, we introduce NGI-Attack, incorporating Example Backtracking and Multiplex Mask strategies to exploit this gradient information and enhance transferability. Specifically, we first adopt Example Backtracking to accumulate Neighbourhood Gradient Information as the initial momentum term. Then, we utilize Multiplex Mask to form a multi-way attack strategy that forces the network to focus on non-discriminative regions, which can obtain richer gradient information during only a few iterations. Extensive experiments demonstrate that our approach significantly enhances adversarial transferability. Especially, when attacking numerous defense models, we achieve an average attack success rate of 95.2\%. Notably, our method can seamlessly integrate with any off-the-shelf algorithm, enhancing their attack performance without incurring extra time costs.
 

  

\end{abstract}

\maketitle

\thispagestyle{empty}

\section{Introduction}
\label{sec:intro}

Deep neural networks (DNNs) have made significant progress~\citep{Arash1, Arash2, Arash3, Arash4, Arash5, Arash6} and are widely applied in security-sensitive applications, such as autonomous driving~\citep{yang2023aide,hong2023lvos, HOANG2024104031}, face anti-spoofing~\citep{10448479,ZhengLWWMLDW24}, and face recognition~\citep{wang2018cosface}. However, recent work~\citep{fgsm,pgd,chen2022shape,chen2022towards,huang2022cmua,chen2023devopatch,jiang2023efficient, Khaleel_Habeeb_Alnabulsi_2024,hybridattack} has revealed that DNNs are highly susceptible and vulnerable to adversarial examples, which introduce imperceptible perturbations to the input and ultimately cause the model to produce erroneous predictions.

Adversarial attacks are typically categorized into white-box and black-box settings. In the white-box settings, gradient-based attacks such as Projected Gradient Descent (PGD)~\citep{pgd} exhibit remarkably high attack success rates on targeted models, capable of accessing all information (\textit{e.g.}, architecture or gradients) of the targeted model. However, in real-world scenarios, attackers often need to generate adversarial examples under black-box settings, as they are unaware of any information of the target model~\citep{optattack,sign-opt,QEBA}. From this perspective, adversarial transferability is crucial, enabling adversaries to exploit surrogate models to generate adversarial examples and transfer them to target models to perform attacks.
Therefore, the generation of highly transferable adversarial examples~\citep{mi,xie2019improving,sini,wang2021admix,dong2019evading, NEURIPS2022_c0f9419c, chen2023aca} has gradually become a hot topic in the community of adversarial examples.

Lin et al. \cite{sini} regard the generation of adversarial examples as an optimization process, and the white-box attack induces poor transferability by causing overfitting of the targeted model. Numerous works have been proposed to alleviate overfitting by enhancing adversarial transferability, such as advanced optimization~\citep{mi,sini,DBLP:journals/corr/abs-1908-06281,DBLP:conf/cvpr/Wang021}, input transformation~\citep{xie2019improving,wang2021admix,dong2019evading,DBLP:journals/corr/abs-1908-06281, SIA, wang2024boosting}, and model augmentation~\citep{ssa}. However, a significant disparity in attack performance persists between the transfer and the optimal white-box settings, necessitating further endeavors to enhance transferability.

Upon recalling these methods, advanced optimization is the most fundamental strategy, \textit{e.g.} MI-FGSM~\citep{mi}, which accumulates gradients to become momentum and leverages momentum to aid the optimization of adversarial examples towards escaping local optima. 
Numerous optimization-based methods have concentrated on refining gradient accumulation and momentum application~\citep{DBLP:journals/corr/abs-1908-06281,DBLP:conf/cvpr/Wang021,NEURIPS2022_c0f9419c}, aiming to fully exploit the capabilities of MI-FGSM.
These enhancements aim to identify optimal strategies for gradient information and enhance adversarial transferability.
However, they exhibit limited transferability because they do not distinguish the transferability differences of gradient information across iterations, instead applying it uniformly in each iteration.
Another strategy, input transformation-based methods, seeks to mitigate attack overfitting by boosting the transferability of adversarial examples through various input augmentation strategies. While combining gradient optimization and input transformation approaches has been shown to significantly increase transferability~\citep{DBLP:journals/corr/abs-1908-06281}, there remains a notable gap in achieving satisfactory performance, especially when attacking numerous defense models.

To tackle the issue above, we propose a novel attack based on Neighbourhood Gradient Information, termed NGI-Attack, incorporating two key strategies: Example Backtracking and Multiplex Mask.
From the perspective of advanced optimization, {we observe that the gradient information generated from the initial iteration, when the adversarial examples are still around the clean images, can offer high transferability. }
We introduce this novel concept as \textit{Neighbourhood Gradient Information (NGI)}.
To demonstrate its effectiveness, we perform a preliminary analysis by incorporating additional iterations before regular attacks to accumulate this specific gradient information via momentum, as detailed in Section~\ref{sec: motivation}.
Utilizing the Example Backtracking method, we collect this gradient information directly.
However, when the number of iterations for the attack is fixed, allocating some of these iterations to collect Neighbourhood Gradient Information consequently necessitates a larger step size to ensure the attack reaches its maximum perturbation magnitude.
Nonetheless, simply increasing the step size results in poor transferability on adversarially trained models. 
To address this, we introduce an input transformation approach, specifically the Multiplex Mask strategy, which forms a multi-way attack.
This strategy employs two distinct branches: one applies MaskProcess images to direct the network’s focus to non-discriminative regions, while the other employs clean images. By incorporating two pathways, we can obtain richer gradient information and reach enough perturbation magnitude.
Furthermore, our method is plug-and-play and requires no additional time overhead, making it easy to integrate with existing strategies. Our contributions are followed as:

\begin{itemize}
\item We validate the hypothesis that 
the Neighbourhood Gradient Information
offers high transferability.

\item 
We propose a novel NGI-Attack that incorporates two key strategies: Example Backtracking and Multiplex Mask. This attack leverages the Neighbourhood Gradient Information and is effective against both normally and adversarially trained models.

\item  Extensive experiments show that our method achieves state-of-the-art performance. In particular, we outperform existing methods by margins of 2.6\% to 19.3\% under normally trained models and 2.4\% to 13.5\% under adversarially trained models. Furthermore, we attain an average success rate of 95.2\% against defense models.
\end{itemize}

The rest of this paper is organized as follows: In Section~\ref{sec:related}, we commence with an overview of both traditional and recent transfer-based black-box attack strategies, followed by a brief review of leading defensive mechanisms. Section \ref{sec:method} delves into the concept of Neighborhood Gradient Information, elucidating its significance in enhancing attack efficacy. Herein, we also introduce our novel strategies and provide the corresponding algorithmic framework.  In Section \ref{sec:exp}, we present experimental results of black-box attacks across various settings, including single-model, multi-model, and defense models. To further analyze our approach, we conduct ablation studies and offer corresponding analysis of the experimental findings. Finally, in Section \ref{sec:conclusion}, we conclude our research and discuss its border impact.

\section{Related Work}
\label{sec:related}
Since adversarial examples are found to exploit DNNs vulnerabilities \citep{L_BFGS}, many adversarial attack methods have been proposed, including white-box and black-box attacks. Unlike white-box attacks, black-box attacks cannot directly access the model's structure or gradient information. Adversarial examples under black-box attacks are mainly generated via (1) Query-based attacks using iterative queries to the target model. (2) Transfer-based attacks applying a surrogate model and then transferring to the target models to perform the attack. In this work, we focus on enhancing the transferability of adversarial examples under black-box attack. A general introduction to transfer attacks is provided as follows. Concurrently, we furnish an overview of recent advancements in adversarial defense.

\subsection{Optimization-based Attacks.} 
These methods primarily aim to effectively exploit gradient information to enhance the transferability of adversarial examples.

\noindent\textbf{Fast Gradient Sign Method (FGSM).} FGSM~\citep{fgsm},
the pioneering algorithm that uses gradient optimization techniques, generates adversarial examples by maximizing the loss function in a single update step:
\begin{equation}
x^{adv} = x + \epsilon \cdot \text{sign}\left( \nabla_x J(x, y; \theta) \right),
\end{equation}
where $\nabla_x J(x, y; \theta)$ denotes the gradient of the loss function with respect to the input $x$, given model parameters $\theta$. The $\text{sign}$ function ensures that the perturbation is applied in the direction of the steepest ascent in the loss landscape, and the $\epsilon$ controls the magnitude of the perturbation.

\noindent\textbf{Iterative Fast Gradient Sign Method (I-FGSM).}
I-FGSM \citep{ifgsm} introduces a small step size $\alpha$, extending FGSM to an iterative version, thereby improving white-box attack performance: 
\begin{equation}
    x_{0}^{adv} = x, \quad x_{t+1}^{adv} = x_{t}^{adv} + \alpha \cdot \text{sign}(\nabla_{x_{t}^{adv}} J(x_{t}^{adv}, y; \theta)).
\end{equation}

\noindent\textbf{Momentum Iterative Fast Gradient Sign Method (MI-FGSM).}
To escape the suboptimal local solution of the surrogate model and to boost the effect of black-box attacks, \cite{dong2017discovering} introduce the momentum term and integrate it into I-FGSM to improve transferability:
\begin{equation}
    g^{t+1} = \mu \cdot g^{t} + \frac{\nabla_{x_{t}^{adv}} J(x_{t}^{adv}, y; \theta)}{\| \nabla_{x_{t}^{adv}} J(x_{t}^{adv}, y; \theta) \|_1}, \quad x_{t+1}^{adv} = x_{t}^{adv} + \alpha \cdot \text{sign}(g^{t+1}),
\end{equation}
where $\mu$ is the decay factor and the initialization of $g^{t}$ is 0.

\noindent\textbf{Nesterov Iterative Fast Gradient Sign Method (NI-FGSM).}
\cite{DBLP:journals/corr/abs-1908-06281} employ the Nesterov accelerated gradient to implement a lookahead in each iteration, thereby circumventing local optimal solutions:
\begin{equation}  
x_{t}^{nes} = x_{t}^{adv} + \alpha \cdot \mu \cdot g_{t}, \quad
g^{t+1} = \mu \cdot g^{t} + \frac{\nabla_{x_{t}^{nes}} J(x_{t}^{nes}, y; \theta)}{\| \nabla_{x_{t}^{nes}} J(x_{t}^{nes}, y; \theta) \|_1},
\quad
x_{t+1}^{adv} = x_{t}^{adv} + \alpha \cdot \text{sign}(g^{t+1}).
\end{equation}

\noindent\textbf{Variance Tuning Gradient-based Attacks (VMI-FGSM).}
In addition, \cite{DBLP:conf/cvpr/Wang021} considers the gradient variance, adjusting the current gradient by its variance to ensure stable updates:
\begin{equation}
g^{t+1} = \mu \cdot g^{t} + \frac{\nabla_{x_{t}^{adv}} J(x_{t}^{adv}, y; \theta) + v_{t}}{\| \nabla_{x_{t}^{adv}} J(x_{t}^{adv}, y; \theta) + v_{t} \|_1}, \quad
v_{t+1} = V(x_{t}^{adv}), \quad
V(x) = \frac{1}{N} \sum_{i=1}^{N} \left( \nabla_{x_i} J(x_i, y; \theta) - \nabla_x J(x, y; \theta) \right).   
\end{equation}

\noindent\textbf{Reverse Adversarial Perturbation (RAP).}
Unlike traditional approaches that focus on a single adversarial point's loss, RAP~\citep{NEURIPS2022_c0f9419c} aims to locate an adversarial example in a region of consistently low loss. Nonetheless, RAP necessitates a greater number of iterative steps to achieve optimal transferability.

\subsection{Input transformation-based Attacks.}
These methods employ an input transformation strategy, which enhances the transferability of adversarial examples by implementing multiple input transformations. 

\noindent\textbf{Diverse Input Method (DI-FGSM).} 
DI-FGSM \citep{xie2019improving} stochastically resizes and pads the input image with a specified probability. The stochastic transformation function $T (X_{t}^{adv} ; p)$ is
\begin{equation}
T (X_{t}^{adv} ; p) = 
\begin{cases} 
T (X_{t}^{adv}) & \text{with probability } p, \\
X_{t}^{adv} & \text{with probability } 1 - p.
\end{cases}
\end{equation}

\noindent\textbf{Translation-Invariant Method (TI-FGSM).}
TI-FGSM \citep{dong2019evading} improves transferability by applying convolutional transformations to gradients with a Gaussian kernel, approximating gradients across an image collection. 
\begin{equation}  
x_{t+1}^{adv} = x_{t}^{adv} + \alpha \cdot \text{sign}(W * \nabla_x J(x_{t}^{adv}, y)).
\end{equation}
where $W$ is the pre-defined kernel.

\noindent\textbf{Scale-Invariant Method (SI-FGSM).}
SI-FGSM \citep{DBLP:journals/corr/abs-1908-06281} leverages the scaling invariance of images by calculating gradients across various scaling factors.

\noindent\textbf{Admix.}
Admix~\citep{wang2021admix} further enhances input images by adding a portion of other images, retaining original labels, and computing gradients of the mixed images:
\begin{equation}  
 \tilde{x} = \gamma \cdot (x + \eta \cdot x'), \quad
\overline{g}_{t+1} = \frac{1}{m_1 \cdot m_2} \sum_{x' \in X'} \sum_{i=0}^{m_1-1} \nabla_{x_t^{adv}} J(\gamma_i \cdot (x_t^{adv} + \eta \cdot x'), y; \theta), \quad
g_{t+1} = \mu \cdot g_t + \frac{\overline{g}_{t+1}}{\|\overline{g}_{t+1}\|_1},
\end{equation}
where $x'$ denotes an image randomly picked from other categories, $\gamma$ and $\eta$ control the proportions of the original image and the mixed image respectively, $m_1$ is the number of admix operations per each $x'$, and $X'$ denotes the set of $m_2$ randomly sampled images from other categories.

\noindent\textbf{SIA.}
Structure Invariant Transformation (SIA)~\citep{SIA} enhances transferability by applying diverse input transformations to localized image blocks for gradient computation.

\noindent\textbf{BSR.}
\cite{wang2024boosting} introduce Block Shuffle and Rotation (BSR) to enhance adversarial transferability. BSR divides the input image into multiple blocks, then randomly shuffles and rotates these blocks to disrupt the intrinsic relationships within the image, demonstrating superior performance in black-box attack settings.

\subsection{Model Augmentation Attacks.}
\noindent\textbf{Spectral Simulation Attack (SSA).}
SSA~\citep{ssa} concentrates on the frequency domain, employing spectral transformations on the inputs to simulate model enhancement methods, thereby boosting the transferability of adversarial examples.  However, this method requires a considerable amount of time.

\subsection{Adversarial Defense}
In this section, we offer a brief introduction to adversarial defenses. Various defense methods have been proposed to mitigate the threat posed by adversarial examples. Adversarial training~\citep{Florian2018Ensemble}, which incorporates adversarial examples into training data, is widely considered one of the most effective approaches to enhance model robustness. However, these methods often entail a significant computational overhead. Another category of methods primarily focuses on efficiently purifying adversarial examples. Specifically, \cite{DBLP:conf/cvpr/LiaoLDPH018} propose a High-level Representation Guided Denoiser (HGD). \cite{xie2017mitigating} introduce random resizing and padding on the input (R\&P). Feature Distillation (FD)~\citep{liu2019feature} put forward a JPEG-based defensive compression framework. Comdefend ~\citep{DBLP:conf/cvpr/JiaWCF19} suggest an end-to-end compression model. \cite{DBLP:conf/icml/CohenRK19} employ a technique known as Randomized Smoothing (RS). Meanwhile, \cite{DBLP:conf/ndss/Xu0Q18} introduce bit reduction (Bit-Red) for the detection of adversarial examples. 
The Neural Representation Purifier (NRP) model by \cite{DBLP:conf/cvpr/NaseerKHKP20} employs a self-supervised adversarial training approach in the input space, focusing on cleansing adversarially perturbed images.
Recently, diffusion models have emerged as powerful generative models~\citep{ho2020denoising,song2021scorebased}. Nie \textit{et al.}~\citep{nie2022DiffPure} introduce DiffPure, which leverages pre-trained diffusion models for adversarial purification, and demonstrate its power.

\section{Method}
\label{sec:method}
In this section, we provide a comprehensive overview of our method. Initially, we observe a pronounced transferability of gradient information around the clean images. Stemming from this observation, we introduce a novel concept termed Neighbourhood Gradient Information. To harness the full potential of this gradient information, we propose the NGI-Attack, incorporating two key strategies: Example Backtracking (EB) and Multiplex Mask (MM).
In particular, EB allows us to accumulate such gradients via a momentum term.
However, given the finite number of iterations, accumulating Neighborhood Gradient Information consumes valuable iteration time. 
Consequently, increasing the iteration step size is crucial to reaching maximum perturbation magnitude within fewer iterations.
Further,
to induce significant perturbation without overfitting to the attack,
we apply the MM strategy to form a multi-way attack.

\begin{figure*}[!t]
\centering
\subfloat[Additional attack rounds are allocated to collect gradient information, with the attack effect gradually increasing and then stabilizing.]
{
  \includegraphics[width=0.32\textwidth]{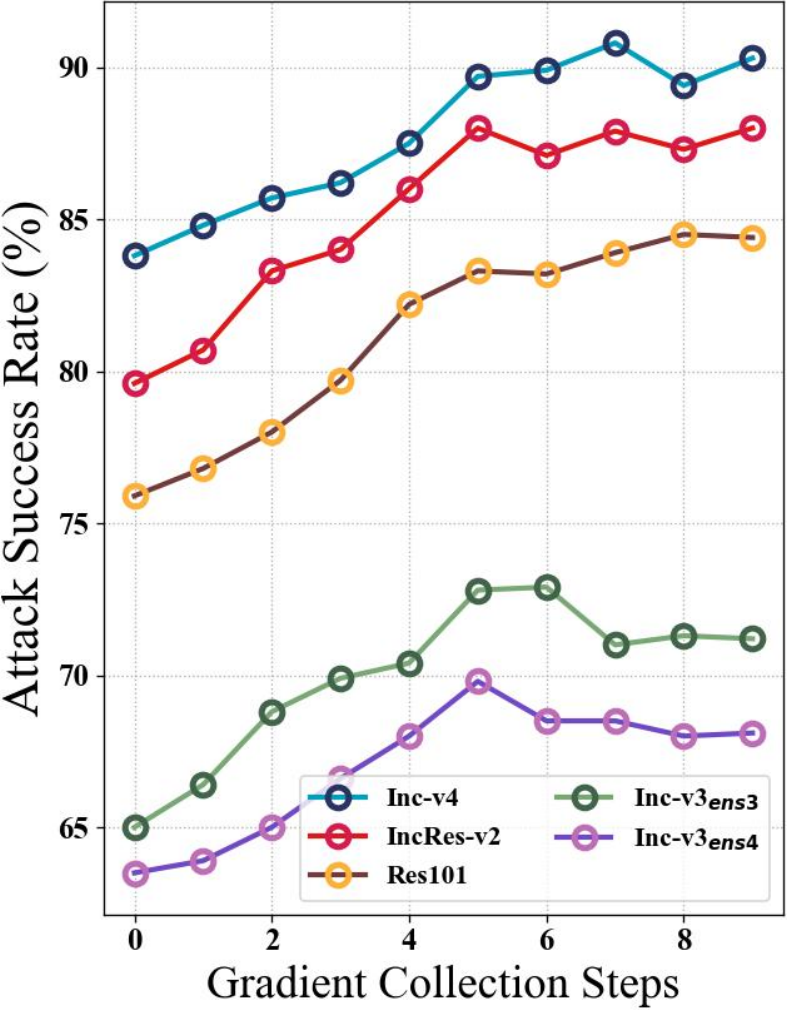}
  \label{fig:toyexp1}
}
\hspace{1em}
\subfloat[Directly scaling the step size produces a significant attack degradation on the defense models.]{
  \includegraphics[width=0.58\textwidth]{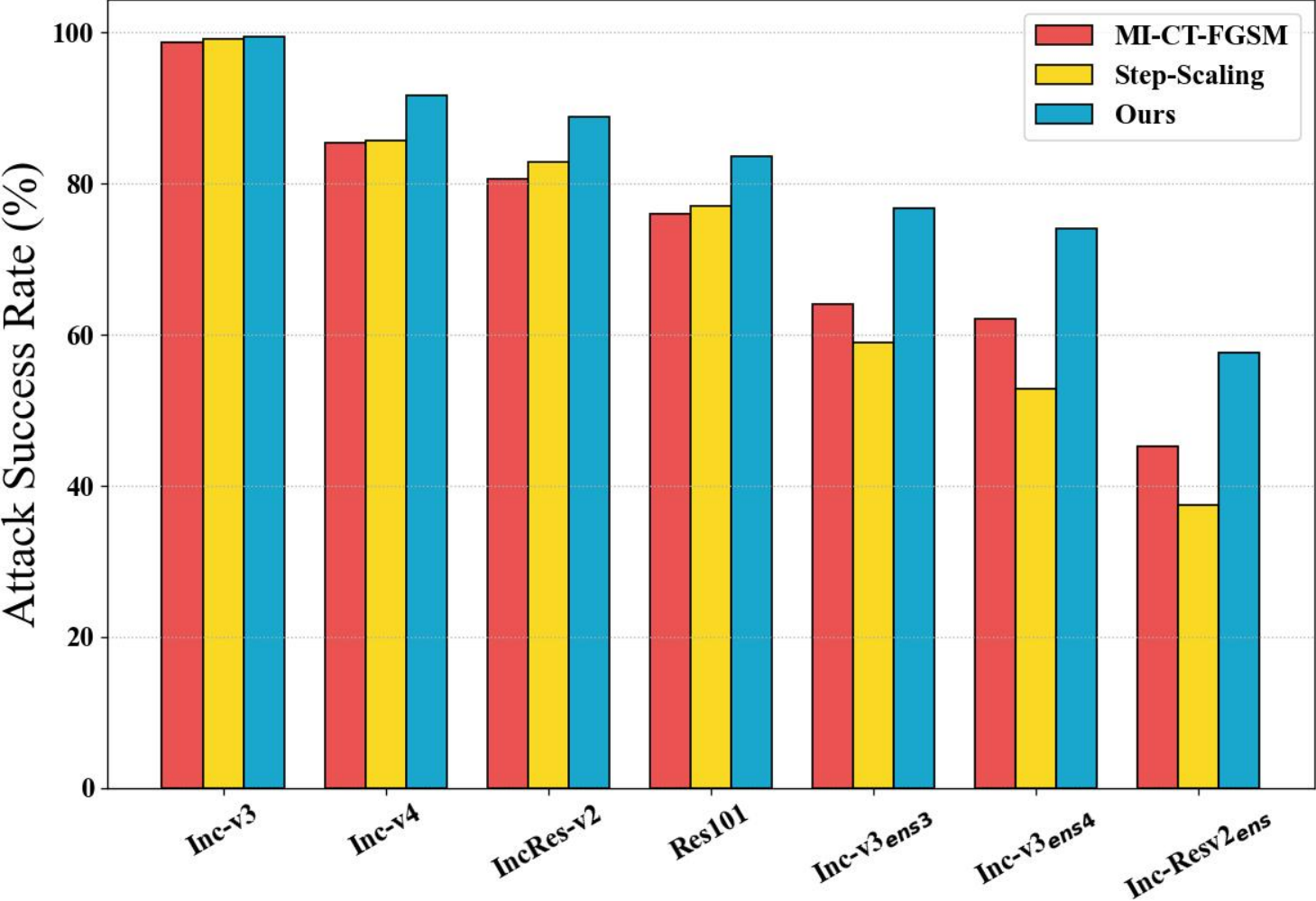}
  \label{fig:toyexp2}
}
\caption{Two toy experiments using Inception-v3~\citep{inceptionv2} as the surrogate model: (a) examines the impact of accumulating Neighborhood Gradient Information, and (b) validates the effects of direct amplification steps on attack performance.}
\label{fig:complete}
\end{figure*}

\subsection{Preliminaries}
Denote $f$: $x$ $\rightarrow$ $y$ as the classification process of a surrogate model, where $x$ is the clean image and $y$ is the corresponding ground-truth label. Let $\mathcal{J}(x, y)$ represent the loss function (\textit{e.g.} the cross-entropy loss). To obtain adversarial examples $x^{adv}$, we need to add a perturbation to $x$ to maximize the loss $\mathcal{J}(x^{adv}, y)$ such that the target model $f$ makes an incorrect classification prediction, \textit{i.e.} $f(x^{adv}) \neq y$ (non-targeted attack). Following ~\citep{xie2019improving,dong2019evading,DBLP:journals/corr/abs-1908-06281,dong2017discovering,DBLP:conf/cvpr/Wang021}, we adopt the $L_\infty$ norm, whereby $\Vert x - x^{adv} \Vert _{\infty} \leq \epsilon$, with $\epsilon$ serving as the upper bound on the constrained perturbation to ensure attack imperceptibility. In the black-box transfer attack, where direct access to the target model $f$ is not available, a surrogate model must be identified as a target for the attack. So the optimization objective can be summarised as:
\begin{equation}
    \mathop{\mathrm{argmax}}\limits_{x_{adv}}(\mathcal{J}(x^{adv},y)), \quad  s.t. ~ ||x - x^{adv}||_\infty \leq \epsilon.
\end{equation}
\cite{fgsm} first propose the Fast Gradient Sign Method (FGSM), which generates adversarial examples through a single-step attack. Subsequently 
\cite{ifgsm} propose the Iterative Fast Gradient Sign Method (I-FGSM), and most gradient optimization-based transfer attack methods are based on I-FGSM. With a total of $T$ iterations, constrain each attack step to be $\alpha = \epsilon / T$. The attack process can be described as follows: 
\begin{equation}
        x^{adv}_{t+1} = Clip_{\epsilon}(x^{adv}_t + \alpha \cdot  {\rm sign}(\nabla_x \mathcal{J}(x^{adv}_t, y))),
\end{equation}
where $x^{adv}_t$ denotes the adversarial example for iteration $t$ and $Clip_{\epsilon}(\cdot)$ constrains the $\delta$ within $\epsilon$.

\begin{figure*}[t]
\centering
\includegraphics[width=1\textwidth]{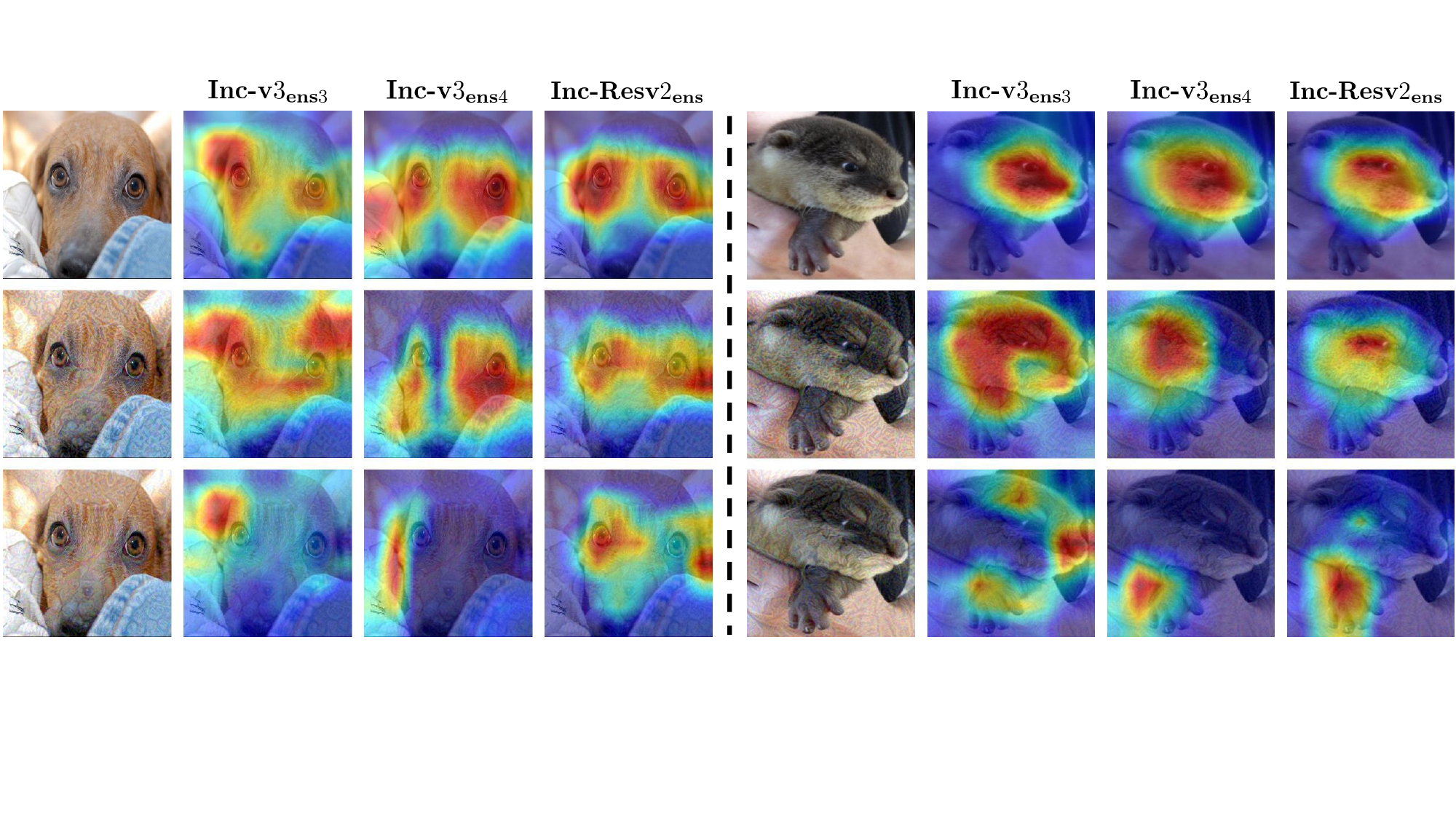}
\caption{Visualization of the attention of adversarial examples under different strategies with Grad-Cam. The top row, the middle row, and the bottom row represent the attention distribution of the clean image, the generated adversarial examples under the scale-up step strategy, and the generated adversarial examples under our strategy, respectively, under different defense models.
{
The results demonstrate that our strategy successfully achieves adversarial attacks by effectively disrupting the network's attention, while the scale-up strategy tends to maintain regions of interest similar to those in the clean images across various defense models.
}
}
\label{fig:fig2}
\end{figure*}

\subsection{Motivation}
\label{sec: motivation}
Existing optimization-based methods in generating adversarial examples often fail to distinguish the transferability differences of gradient information across iterations, applying it uniformly in each iteration. They overlook a crucial phenomenon: as iterations increase, the gradient information may tend to overfit the surrogate model, thereby impairing its transferability to other models. In contrast, we observe that the gradient information generated from the initial iteration, when the adversarial examples are still around the clean images, exhibits superior transferability. We define this high-transferability gradient information, derived around the clean images, as \textit{Neighbourhood Gradient Information}.

To further verify whether the Neighbourhood Gradient Information enhances the attack transferability, we conduct an experiment.
Specifically, we introduce additional rounds of attacks near the clean images before executing attack, and we use momentum to accumulate the Neighbourhood Gradient Information. 
As shown in Figure~\ref{fig:toyexp1}, collecting the Neighbourhood Gradient Information for just one round in advance significantly improves the attack success rate and reaches a high attack success rate at 5 rounds of collection. 
However, as the number of iterations increases, the attack effectiveness tends to stabilize. This indicates that collecting gradient information from later iterations, when the adversarial examples are away from the clean images, has a diminishing impact on improving adversarial transferability.
In conclusion, the perturbations generated around the clean images are indeed more effective for the attack.
{This empirical finding is consistent with recent theoretical analyses showing that early-stage gradients reside in low-frequency, flat regions of the loss landscape—features widely shared across models and thus highly transferable \citep{pmlr-v97-rahaman19a,sagun2017eigenvalues,cohen2021gradient,foret2021sharpnessaware,Yinrobustness,Lidefense,hu2020adversarial}.}

To maximize the utilization of the Neighbourhood Gradient Information without additional time costs, we attempt to increase the step size in each iteration, thereby enhancing the perturbation magnitude. However, we observe that directly increasing the step size may weaken the transferability of adversarial examples.
As illustrated in Figure~\ref{fig:toyexp2}, when the step size is doubled, \textit{i.e.}, step-scaling, there is a slight increase in the attack success rate on normally trained models, highlighting the effectiveness of Neighbourhood Gradient Information. 
However, against adversarially trained models, there is a notable decrease, ranging from 5.2\% to 9.3\%. 
To further elucidate the factors underlying this phenomenon, we employ Grad-Cam~\citep{gradcam} to analyze the behavior of various defense models visually. Figure~\ref{fig:fig2} presents the clean images, adversarial examples generated using the scale-up step strategy, and those produced by our approach, along with their corresponding Grad-Cam heatmaps across different defense models.
{
The results indicate that successful adversarial attacks disrupt the network's attention, as evidenced by significant divergence in Grad-Cam heatmaps from the clean images. Conversely, unsuccessful attacks tend to maintain regions of interest similar to those in the clean images.} 
Our findings reveal that adversarial examples generated by the scale-up strategy exhibit consistent attention patterns across defense models, indicating limited transferability, likely due to the overfitting associated with excessively large attack steps. In contrast, our method effectively disrupts attention across various defense models, thereby demonstrating enhanced transferability.
Therefore, the question arises: Can we find a more straightforward and effective method that improves the attack success rate without incurring the additional time cost associated with accumulating the Neighbourhood Gradient Information?

\begin{figure*}[t]
\centering
\includegraphics[width=1\textwidth]{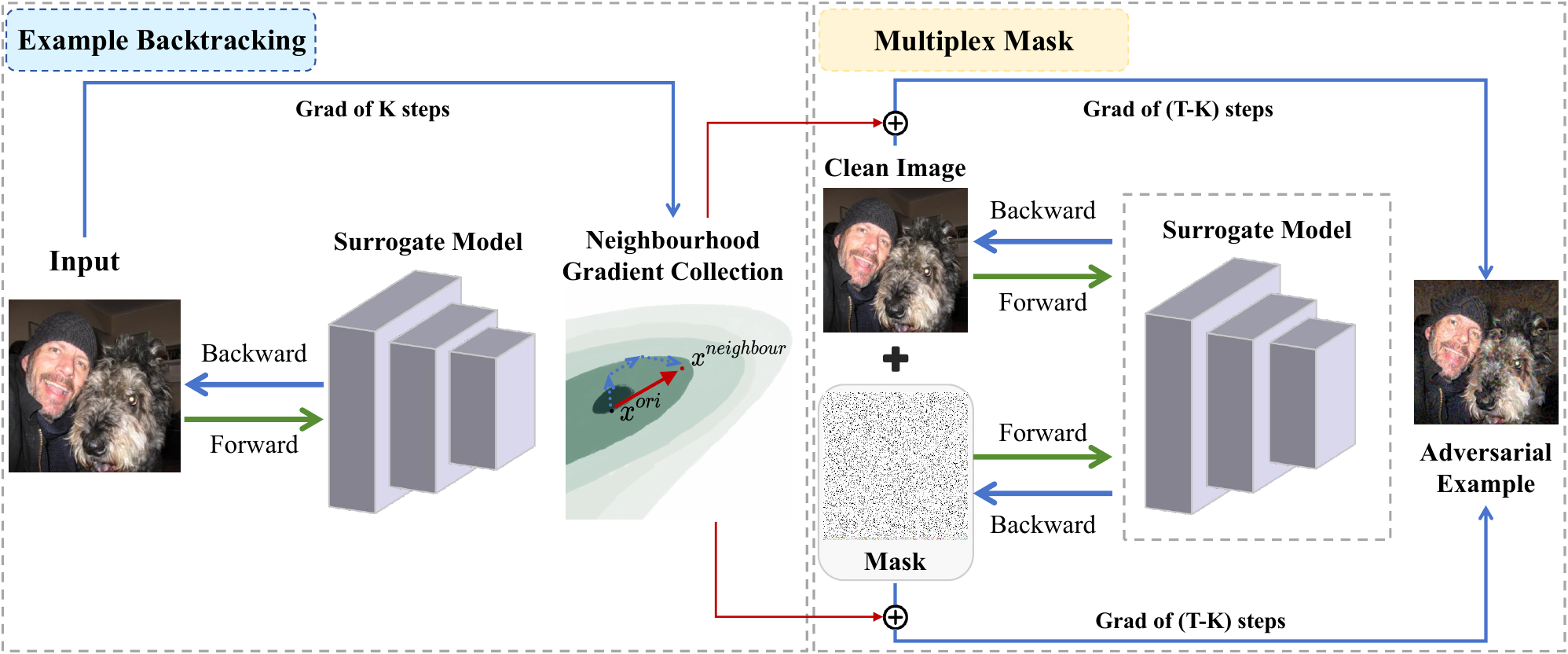}
\caption{
The overall NGI-Attack pipeline: The process begins using the Example Backtracking strategy to collect Neighbourhood Gradient Information. This information is then utilized in the Multiplex Mask strategy, forming a multi-way attack. One pathway applies the MaskProcess operation, while the other directly utilizes the clean image. The gradients of these two pathways are collected independently, guided by the Neighbourhood Gradient Information, ultimately generating a highly transferable adversarial example.
}
\label{fig:pipeline}
\end{figure*}

\subsection{NGI-Attack}
To address the aforementioned issue and effectively leverage the
Neighbourhood Gradient Information, we propose the NGI-Attack, incorporating two key strategies: Example Backtracking (EB) and Multiplex Mask (MM).
The overall NGI-Attack pipeline is depicted in Figure~\ref{fig:pipeline}.

\textbf{Example Backtracking.} 
EB performs a Neighbourhood Gradient Collection, directly accumulating gradient information, which is subsequently utilized to generate adversarial examples. 
Moreover, to enhance the collection and utilization of Neighbourhood Gradient Information, additional processing of the obtained gradient information is necessary when employing I-FGSM. As a result, we directly utilize MI-FGSM as our baseline, which incorporates momentum to collect the gradient information effectively. To mitigate the introduction of additional time costs, we adopt a two-stage approach in the overall attack process.
Specifically, we dedicate the initial $K$ rounds to Neighbourhood Gradient Collection, followed by the remaining $(T - K)$ rounds, where we conduct substantial attacks to generate adversarial examples.
Thus, the Neighbourhood Gradient Collection process during the attack is as follows: 
\begin{equation}
    g_{t+1} = \mu \cdot g_t + \frac{\nabla_x \mathcal{J}(x^{adv}_t, y)}{||\nabla_x \mathcal{J}(x^{adv}_t, y)||_1}, \quad
    x^{adv}_{t+1} = \text{Clip}_{\epsilon} \left(x^{adv}_t + \alpha \cdot \text{sign}(g_{t+1})\right),
\label{eq1}
\end{equation}
where $g_t$ gathers gradients of previous $t$ iterations with an initial value of $g_0$ = 0 and $\mu$ is the decay factor of momentum term. 
Subsequently, Example Backtracking performs a simple reset of the example back to the clean image after finishing $K$ rounds of Neighbourhood Gradient Collection.
That is, when the number of iteration rounds $t$ in the iterative attack is equal to $K$, reset $x^{adv}_K$ = $x^{adv}_0$.
To formalize the gradient accumulation during this phase, we define the Neighbourhood Gradient Information (NGI) as:
\begin{equation}
        g_{\text{NGI}} = g_K = \sum_{t=1}^{K} \mu^{K - t + 1} \cdot \frac{\nabla_x \mathcal{J}(x_{t-1}^{adv}, y)}{\left\| \nabla_x \mathcal{J}(x_{t-1}^{adv}, y) \right\|_1}.
    \label{eq:ngi}
\end{equation}
This $g_{\text{NGI}}$ then serves as the initialized momentum term for the subsequent $(T-K)$ attack rounds.
But direct use of Example Backtracking inevitably poses a problem: to ensure that the attack reaches the maximum perturbation limit without extra time costs, it is necessary to employ a larger step size in the latter $(T - K)$ rounds.  However, as indicated by the previous analysis, directly increasing the step size may lead to overfitting of the attack.
Therefore, obtaining more effective information about the attack gradient within the same iteration rounds becomes a crucial factor in improving the attack success rate.

\begin{figure*}[t]
\centering
\includegraphics[width=1\textwidth]{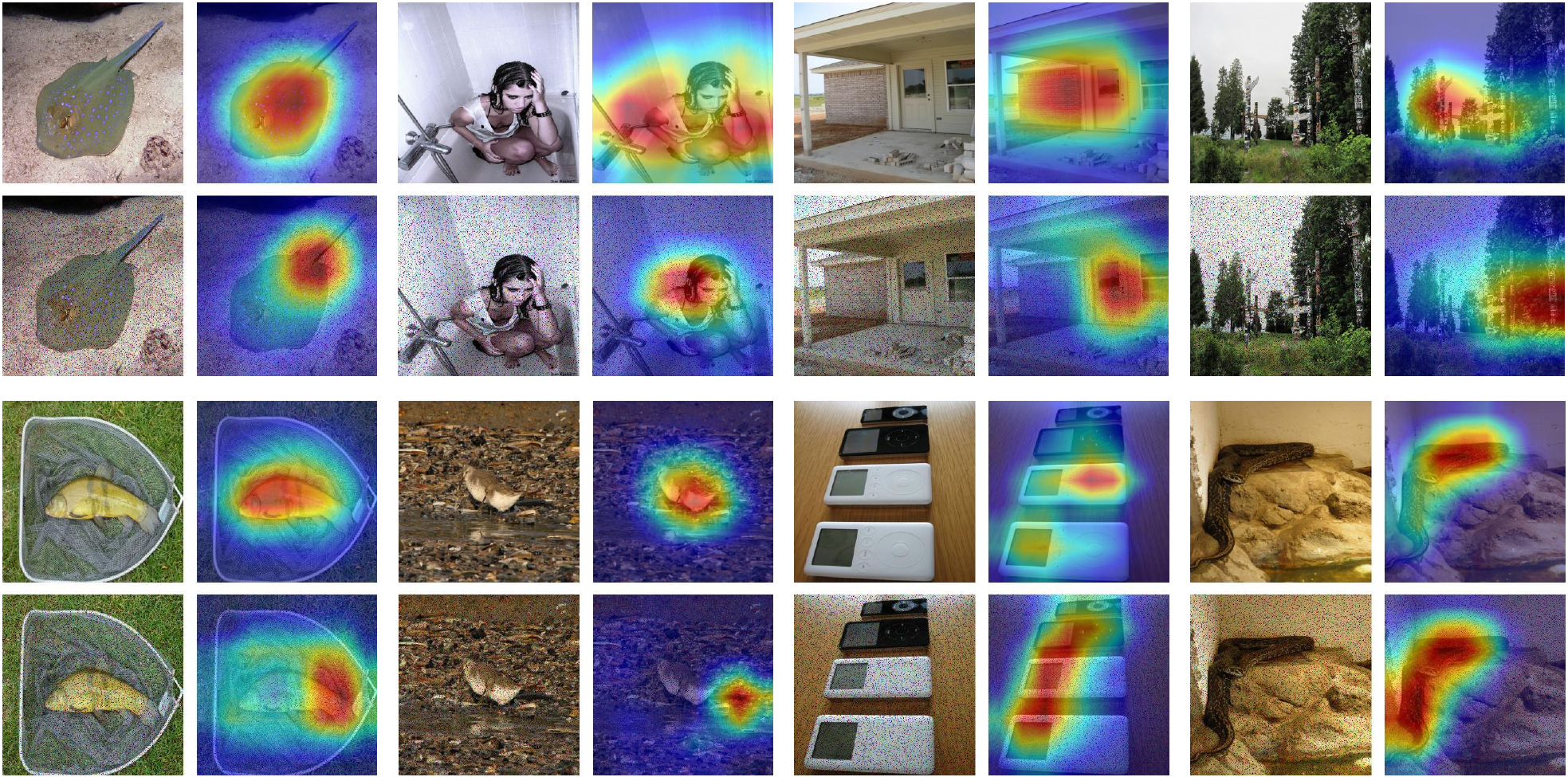}
\caption{Visualization of the attention of MaskProcess images with towards Inception-v3~\citep{inceptionv2}. Each set comprises two rows; the first represents the clean image and its attention visualization, while the second represents the MaskProcess image with $P$ of 0.9 and its attention visualization.
The results show that the MaskProcess image can successfully shift the network's attention towards non-discriminative regions.
}
\label{fig:app_c}
\vspace{-0.5cm}
\end{figure*}

\textbf{Multiplex Mask.}
Direct use of increasing step size leads to overfitting of the attack process.
We attribute this phenomenon to the repetitive use of the same gradient information.
A more sophisticated approach is to gather diverse gradient information. This allows us to effectively split the larger step size, rather than linearly scaling it, which in turn suppresses the aforementioned overfitting issue.
The defense model turns out to have a more expansive attention region for the target compared to the normally trained model~\citep{dong2019evading}, 
and according to Figure~\ref{fig:toyexp2}, when images are subjected to attacks with excessively large step sizes, they may weaken transferability towards adversarially trained models.
To counter this, we propose to augment the input with a mask. This makes the model focus on more non-discriminative regions, effectively expanding the attention region. Subsequently, it enables a more comprehensive gradient information extraction and a more thorough feature destruction.

To visually underscore this phenomenon, we present the MaskProcess in Figure \ref{fig:app_c}. The figure shows that the network's attention shifts towards the non-discriminative regions after applying the masking strategy.
By using the mask-enhanced image,
we capture new gradient information across different feature regions, thereby enhancing the adversarial transferability:
\begin{equation}
    x_{t}^{mask} = \operatorname{MaskProcess}(x_{t}^{adv}, P), \quad
    g_{t+1}^{mask} = \mu \cdot g_t^{mask} + \frac{\nabla_x \mathcal{J}(x^{mask}_t, y)}{||\nabla_x \mathcal{J}(x^{mask}_t, y)||_1},
\label{eq2}
\end{equation}

\textit{where $\operatorname{MaskProcess(x, \mathit{P})}$ represents a masking operation on the input $x$ with probability $P$, \textit{i.e.} for each pixel of the input, the pixel is kept constant with probability $P$ and conversely becomes 0.} 

During the attack process, we form a multi-way attack. One pathway involves applying the $MaskProcess$ operation, while the other pathway maintains the inputs unchanged. The gradients of these two pathways are collected independently, enabling distinct gradient information for each pathway. However, during the iterative generation of adversarial examples, the momentum information from both pathways is utilized to update the momentum. As a consequence, the iterative attack process deviates from the original Equation~\ref{eq1} and is transformed into the following formulation:
\begin{equation}
    x_{t+1}^{adv} = Clip(x_t^{adv} + \alpha \cdot sign(g_{t+1}) + \alpha \cdot sign(g_{t+1}^{mask})).
\end{equation}
This approach effectively addresses the issue of directly increasing step size and facilitates the formation of a more comprehensive disruption of features. 
Overall,  we can effectively gather additional information around the clean images without overfitting the attack process. 
The entire process of NGI-Attack is outlined in Algorithm~\ref{alg:fi}.

\renewcommand{\algorithmicrequire}{\textbf{Input:}}
\renewcommand{\algorithmicensure}{\textbf{Output:}}
\newcommand{\algorithmicinitialize}{\textbf{Initialize:}}
\newcommand{\INITIALIZE}[1]{\item[\algorithmicinitialize] #1}

\begin{algorithm}
\caption{NGI-Attack}\label{alg:fi}
\begin{algorithmic}[1]
    \REQUIRE A classifier $f$ with parameters $\theta$; loss function $J$; a clean image $x$ with ground-truth label $y$; iterations $T$; maximum perturbation $\epsilon$; initial state backing stage $K$; mask probability $P$;
    \ENSURE An adversarial example $x^{adv}_T$;
    \INITIALIZE{$g_0 = 0; g_0^{mask} = 0; x_0^{adv} = x; \alpha = \epsilon / T$;}\\
    \FOR{$t = 0$ \textbf{to} $T-1$}
        \STATE $x_t^{mask} = \text{MaskProcess}(x_t^{adv}, P);$ 
        \STATE $g_{t+1} = \mu \cdot g_t + \frac{\nabla \mathcal{J}(x_t^{adv}, y)}{\|\nabla \mathcal{J}(x_t^{adv}, y)\|_1};$ 
        \STATE $g_{t+1}^{mask} = \mu \cdot g_t^{mask} + \frac{\nabla \mathcal{J}(x_t^{mask}, y)}{\|\nabla \mathcal{J}(x_t^{mask}, y)\|_1};$
        \IF{$t = K$}
            \STATE $x_t^{adv} = x;$ 
        \ENDIF
        \STATE $x_{t+1}^{adv} = \text{Clip}_\epsilon(x_t^{adv} + \alpha \cdot \text{sign}(g_{t+1}) + \alpha \cdot \text{sign}(g_{t+1}^{mask}));$
    \ENDFOR
\end{algorithmic}
\end{algorithm}

\textbf{Time Complexity Analysis.}
Our NGI-Attack method enhances adversarial transferability without introducing additional time costs. During the first $K$ iterations, we apply the EB strategy to gather Neighbourhood Gradient Information. For the subsequent $(T-K)$ iterations, we employ the MM strategy, a multi-way attack approach that processes two inputs simultaneously. These operations are completed within an $O(1)$ time complexity and do not incur additional computational overhead. Therefore, despite involving more complex operations, the overall time cost of our method does not increase, maintaining the total number of iterations at $T$. This aligns with the time complexity of previous iterative attack methods ~\citep{mi,xie2019improving,dong2019evading}.

To further validate the theoretical analysis, we conduct an empirical runtime evaluation using Inception-v3 as the surrogate model. All methods are implemented under the same settings (batch size = 10) on an NVIDIA Tesla V100 (32GB). The results show that NGI-Attack only incurs a moderate increase in runtime compared to MI-FGSM and PGD (1.38s vs. 0.87s/0.88s per batch), while being significantly faster than VMI-FGSM (11.63s per batch). These results confirm that our method improves adversarial transferability without introducing noticeable runtime overhead.

\section{Experiments}
\label{sec:exp}

To validate the effectiveness of our proposed NGI-Attack, we conduct comprehensive experiments on the standard ImageNet dataset~\citep{imagenet}. This section initiates with a thorough outline of our experimental setup. Subsequently, we compare our approach with competitive baselines across varied experimental conditions and provide a quantitative analysis of attack performance on nine advanced defense models. Results consistently underscore that our method significantly enhances transferability across diverse scenarios. For further investigation, we also examine the effects of different proposed modules and the influence of hyperparameters, namely the Example Backtracking step $K$ and the Multiplex Mask probability $P$.

\subsection{Experiment Setup}

\textbf{Datasets.} Following~\citep{DBLP:conf/iclr/KurakinGB17, DBLP:conf/cvpr/Wang021, ssa, dong2017discovering, DBLP:journals/corr/abs-1908-06281}, we utilize the dataset employed in the NIPS 2017 adversarial competition\footnote{\url{https://github.com/cleverhans-lab/cleverhans/tree/master/cleverhans_v3.1.0/examples/nips17_adversarial_competition/dataset}}, consisting of 1000 images randomly picked from the ILSVRC 2012 validation set. These images are ensured to remain within the 1000 categories and to be accurately classified by the test models.

\textbf{Models.} 
In our study, we undertake a comprehensive evaluation of diverse models. We consider four normally trained models, comprising Inception-v3 (Inc-v3)~\citep{inceptionv2}, Inception-v4 (Inc-v4)~\citep{szegedy2017inception}, Inception-Resnet-v2 (IncRes-v2), and Resnet-v2-101 (Res-101)~\citep{DBLP:conf/eccv/HeZRS16}. Additionally, we evaluate two recent transformer-based models, namely Vision Transformer (ViT)~\citep{vit} and Swin Transformer (Swin-b)~\citep{swint}. Furthermore, we assess three adversarially trained models, which are ens3-adv-Inception-v3 (Inc-v3$_{ens3}$), ens4-adv-Inception-v3 (Inc-v3$_{ens4}$) and ens-adv-Inception-ResNet-v2 (IncRes-v2$_{ens}$)~\citep{Florian2018Ensemble}. 
Beyond the models, our study also delves into nine advanced defense methods renowned for their robustness against black-box attacks on the ImageNet dataset, specifically HGD~\citep{DBLP:conf/cvpr/LiaoLDPH018}, R{\&}P~\citep{xie2017mitigating}, NIPS-r3\footnote{\url{https://github.com/anlthms/nips-2017/tree/master/mmd}}, Bit-Red~\citep{DBLP:conf/ndss/Xu0Q18}, JPEG~\citep{guo2017countering}, FD~\citep{liu2019feature}, ComDefend~\citep{DBLP:conf/cvpr/JiaWCF19}, NRP~\citep{DBLP:conf/cvpr/NaseerKHKP20}, and RS~\citep{DBLP:conf/icml/CohenRK19}. We apply the official models for the first three, while for the remaining models, we test their effectiveness using Inc-v3$_{ens3}$. 

\textbf{Baselines.}
We select three typical optimization-based attacks, specifically MI-FGSM~\citep{mi}, NI-FGSM~\citep{sini}, and VMI-FGSM~\citep{DBLP:conf/cvpr/Wang021}, as our baseline comparisons.
For input transformation-based attacks, we opt for DI-FGSM~\citep{xie2019improving}, TI-FGSM~\citep{dong2019evading}, and SI-FGSM ~\citep{DBLP:journals/corr/abs-1908-06281} as our baselines. 
We also include SGM~\citep{sgm} as a baseline.
Moreover, to validate the efficacy of our approach, we contrast it with various recent state-of-the-art attack methods, including SI-NI-FGSM, VT-FGSM~\citep{DBLP:conf/cvpr/Wang021}, Admix~\citep{wang2021admix}, SSA~\citep{ssa}, RAP~\citep{NEURIPS2022_c0f9419c}, SIA~\citep{SIA}, BSR~\citep{wang2024boosting}, and combined variants of these methods, such as VT-TI-DIM (a combined version of VT-FGSM, DI-FGSM, and TI-FGSM).

\textbf{Hyperparameters Settings.} 
In all experiments, the maximum perturbation $\epsilon = 16$, the number of iterations $T = 10$, and the iteration step size $\alpha = \epsilon / T = 1.6$. Notably, the RAP experiment differs from this norm, employing an iteration count of 400. For MI-FGSM and NI-FGSM, we set the decay factor $\mu = 1.0$. For input transformation-based attacks, we assign the transformation probability of $0.5$ for DI-FGSM, apply a $7 \times 7$ Gaussian kernel for TI-FGSM, and use a scale copies number of $5$ for SI-FGSM. For VT-FGSM, we set its hyperparameter $\beta = 1.5$, and the number of sampling examples is 20. For Admix, the scale copies number $m_{1} = 5$, the example number $m_{2} = 3$, and the admix ratio $\mu = 0.2$. For SSA, tuning factor $\rho$ is 0.5, and spectrum transformations $N$ is 20. 
For SIA, the splitting number $s = 3$, and the number of transformed images $N = 20$.
For BSR, the image is split into $2 \times 2$ blocks with a maximum rotation angle of $\tau = 24^\circ$.
Our method uses a backtracing step $K$ of 5, and mask pixel probability $P$ is 0.9.

\begin{table*}[t]
\centering
\caption{Attack success rates (\%) of optimization-based attacks on nine normally and adversarially trained models in the single-model setting. The adversarial examples are crafted on Inc-v3, Inc-v4, IncRes-v2, and Res101, respectively. * indicates the white-box model.}
\label{tab:opt}
\scalebox{0.95}{
\begin{tabularx}{\textwidth}{@{}>{\centering\arraybackslash}X|c|*{10}{>{\centering\arraybackslash}X}@{}}
\toprule
Model & Attack & Inc-v3 & Inc-v4 & Inc-Res-v2 & Res-101 & Inc-v3$_{ens3}$ & Inc-v3$_{ens4}$ & Inc-Res-v2$_{ens}$ & ViT & Swin-b & Avg. \\ \midrule
\multirow{6}{*}{\shortstack{Inc\\-v3}} & MI-FGSM & \textbf{100.0} & 45.2 & 42.3 & 36.7 & 13.5 & 12.9 & 6.3 & 7.9 & 9.8 & 30.5 \\
 & NGI-MI-FGSM(ours) & \textbf{100.0} & \textbf{61.6} & \textbf{61.0} & \textbf{50.6} & \textbf{20.0} & \textbf{21.5} & \textbf{11.0} & \textbf{12.7} & \textbf{15.2} & \textbf{39.3} \\ \cmidrule(l){2-12}
 & NI-FGSM & \textbf{100.0} & 49.1 & 47.7 & 39.0 & 14.0 & 13.0 & 6.1 & 8.0 & 11.5 & 32.0 \\
 & NGI-NI-FGSM(ours) & \textbf{100.0} & \textbf{58.8} & \textbf{55.5} & \textbf{46.5} & \textbf{15.6} & \textbf{15.8} & \textbf{7.9} & \textbf{10.4} & \textbf{12.7} & \textbf{35.9} \\ \cmidrule(l){2-12}
 & VMI-FGSM & \textbf{100.0} & 71.6 & 68.8 & 60.3 & 32.4 & 30.8 & 17.7 & 17.8 & 22.4 & 46.9 \\
 & NGI-VMI-FGSM(ours) & \textbf{100.0} & \textbf{82.4} & \textbf{79.0} & \textbf{73.2} & \textbf{46.0} & \textbf{43.6} & \textbf{29.5} & \textbf{23.9} & \textbf{29.4} & \textbf{56.3} \\ \midrule
\multirow{6}{*}{\shortstack{Inc\\-v4}} & MI-FGSM & 55.0 & \textbf{99.7} & 47.5 & 43.0 & 16.2 & 15.8 & 7.6 & 10.2 & 14.6 & 34.4 \\
 & NGI-MI-FGSM(ours) & \textbf{73.6} & 99.4 & \textbf{63.5} & \textbf{54.1} & \textbf{26.3} & \textbf{25.8} & \textbf{16.5} & \textbf{16.5} & \textbf{22.8} & \textbf{44.3} \\ \cmidrule(l){2-12}
 & NI-FGSM & 60.6 & \textbf{100.0} & 50.9 & 43.5 & 17.4 & 14.7 & 7.0 & 10.8 & 15.4 & 35.6 \\
 & NGI-NI-FGSM(ours) & \textbf{70.9} & 99.9 & \textbf{61.4} & \textbf{51.4} & \textbf{18.0} & \textbf{17.4} & \textbf{9.6} & \textbf{13.1} & \textbf{17.9} & \textbf{40.0} \\ \cmidrule(l){2-12}
 & VMI-FGSM & 77.4 & 99.7 & 70.6 & 62.5 & 38.7 & 36.9 & 23.6 & 22.9 & 31.0 & 51.5 \\
 & NGI-VMI-FGSM(ours) & \textbf{86.4} & \textbf{99.8} & \textbf{81.1} & \textbf{73.2} & \textbf{50.9} & \textbf{48.5} & \textbf{37.0} & \textbf{28.6} & \textbf{40.0} & \textbf{60.6} \\ \midrule
\multirow{6}{*}{\shortstack{IncRes\\-v2}} & MI-FGSM & 58.4 & 50.3 & \textbf{97.7} & 45.3 & 22.5 & 16.5 & 12.1 & 10.4 & 13.8 & 36.3 \\
 & NGI-MI-FGSM(ours) & \textbf{76.0} & \textbf{69.9} & \textbf{97.7} & \textbf{64.2} & \textbf{36.0} & \textbf{28.1} & \textbf{24.1} & \textbf{15.7} & \textbf{20.7} & \textbf{48.0} \\ \cmidrule(l){2-12}
 & NI-FGSM & 63.1 & 54.7 & \textbf{99.0} & 45.9 & 21.1 & 16.8 & 11.8 & 10.4 & 13.9 & 37.4 \\
 & NGI-NI-FGSM(ours) & \textbf{74.8} & \textbf{67.9} & 98.6 & \textbf{57.7} & \textbf{25.8} & \textbf{19.8} & \textbf{14.3} & \textbf{12.3} & \textbf{14.8} & \textbf{42.9} \\ \cmidrule(l){2-12}
 & VMI-FGSM & 79.0 & 72.9 & 97.8 & 68.3 & 47.8 & 40.4 & 33.7 & 23.8 & 28.2 & 54.7 \\
 & NGI-VMI-FGSM(ours) & \textbf{84.6} & \textbf{82.7} & \textbf{97.9} & \textbf{76.6} & \textbf{59.5} & \textbf{53.0} & \textbf{51.9} & \textbf{32.2} & \textbf{41.0} & \textbf{64.4} \\ \midrule
\multirow{6}{*}{\shortstack{Res\\-101}} & MI-FGSM & 56.9 & 50.4 & 49.3 & \textbf{99.3} & 24.8 & 21.7 & 13.0 & 11.3 & 13.8 & 37.8 \\
 & NGI-MI-FGSM(ours) & \textbf{71.5} & \textbf{66.4} & \textbf{64.4} & \textbf{99.3} & \textbf{35.5} & \textbf{32.3} & \textbf{22.4} & \textbf{16.8} & \textbf{19.4} & \textbf{47.6} \\ \cmidrule(l){2-12}
 & NI-FGSM & 62.9 & 55.9 & 53.2 & \textbf{99.3} & 25.7 & 22.2 & 12.3 & 12.4 & 15.4 & 39.9 \\
 & NGI-NI-FGSM(ours) & \textbf{70.2} & \textbf{64.6} & \textbf{64.3} & 99.2 & \textbf{30.1} & \textbf{24.6} & \textbf{15.4} & \textbf{14.1} & \textbf{16.4} & \textbf{44.3} \\ \cmidrule(l){2-12}
 & VMI-FGSM & 74.7 & 69.9 & 70.1 & 99.2 & 45.4 & 41.5 & 29.6 & 22.0 & 25.2 & 53.1 \\
 & NGI-VMI-FGSM(ours) & \textbf{83.4} & \textbf{79.2} & \textbf{79.8} & \textbf{99.4} & \textbf{59.0} & \textbf{53.7} & \textbf{44.6} & \textbf{27.0} & \textbf{32.2} & \textbf{62.0} \\ \bottomrule
\end{tabularx}
}
\end{table*}

\begin{table*}[h]
\centering
\caption{Attack success rates (\%) of SGM on nine normally and adversarially trained models in the single-model setting. The adversarial examples are crafted on Res50 and DN121, respectively. * indicates the white-box model.}
\label{tab:sgm}
\scalebox{0.95}{
\begin{tabularx}{\textwidth}{@{}>{\centering\arraybackslash}X|c|*{10}{>{\centering\arraybackslash}X}@{}}
\toprule
Model & Attack & Inc-v3 & Inc-v4 & Inc-Res-v2 & Res-101 & Inc-v3$_{ens3}$ & Inc-v3$_{ens4}$ & Inc-Res-v2$_{ens}$ & ViT & Swin-b & Avg. \\ \midrule
\multirow{2}{*}{Res50} & SGM & 31.0 & 24.4 & 24.6 & 32.9 & 21.9 & 22.9 & 16.8 & 11.2 & 17.2 & 22.5 \\
 & NGI-SGM & \textbf{49.6} & \textbf{40.7} & \textbf{40.0} & \textbf{49.0} & \textbf{37.8} & \textbf{36.6} & \textbf{28.5} & \textbf{19.4} & \textbf{25.5} & \textbf{36.3} \\ \midrule
\multirow{2}{*}{DN121} & SGM & 34.8 & 28.4 & 25.8 & 39.5 & 29.4 & 29.1 & 20.0 & 11.0 & 16.4 & 26.0 \\
 & NGI-SGM & \textbf{59.0} & \textbf{51.1} & \textbf{49.5} & \textbf{61.6} & \textbf{48.7} & \textbf{44.4} & \textbf{36.7} & \textbf{20.5} & \textbf{29.0} & \textbf{44.5} \\ \bottomrule
\end{tabularx}
}
\end{table*}

\subsection{Attack Evaluation}

\subsubsection{Evaluation on Single-model Attacks} 
In this section, we initially investigate the vulnerability of neural networks in the single-model setting. Specifically, we choose a surrogate model to generate adversarial examples and assess their effectiveness across nine diverse target models, including normally trained and adversarially trained CNNs, as well as transformer-based architectures.
We employ the \textit{Attack Success Rate (ASR)} as our evaluation metric, which is the misclassification rate of corresponding models against adversarial examples. A higher value indicates that the adversarial examples exhibit greater transferability.

We compare with the chosen optimization-based and input transformation-based baselines. Moreover, we evaluate our method against recent state-of-the-art methods. The subsequent content elucidates the aforementioned experimental results in detail, respectively.

\textbf{Attack on optimization-based attacks.}
Initially, we conduct attacks using optimization-based baselines, MI-FGSM, NI-FGSM, and VMI-FGSM. When combined with our approach, denoted as NGI-MI-FGSM, NGI-NI-FGSM, and NGI-VMI-FGSM, we assess their enhancement in attack performance over the baselines.

Table \ref{tab:opt} illustrates that our method consistently boosts the ASR across all black-box models while maintaining comparable performance in white-box settings. 
Specifically, NGI increases the average ASR by 8.8\%–11.7\% when applied to MI-FGSM, 3.9\%–5.5\% for NI-FGSM, and up to 9.7\% for VMI-FGSM, with no performance degradation in any scenario.
More notably, under normally trained models, NGI-MI-FGSM achieves substantial improvements in transferability, ranging from 11.1\% to 19.3\%, while on adversarially trained models, the improvement remains consistent within 4.7\%–13.5\%. In addition, NGI also enhances cross-architecture transferability, with an average improvement of approximately 6.0\% on transformer-based models such as ViT and Swin-B. These results collectively highlight the generalizability and superiority of our approach.

\textbf{Attack on SGM.}
To further demonstrate the generality of NGI, we apply it to SGM, which improves adversarial transferability by suppressing gradients from residual blocks and instead emphasizes gradients from skip connections. We choose ResNet-50 and DenseNet-121 as surrogate models, as their architectures feature extensive skip connections, which are essential for the effectiveness of SGM.

As shown in Table~\ref{tab:sgm}, NGI-SGM achieves substantial improvements over the SGM baseline across all target models. Specifically, under the ResNet-50 surrogate, NGI-SGM raises the average ASR by 13.8\%, while under DenseNet-121, the gain reaches 18.5\% . These results further demonstrate the effectiveness of NGI as a general plug-in strategy across diverse attack paradigms.

\begin{table*}[!th]
\centering
\caption{Attack success rates (\%) of nine normally and adversarially trained models employing both optimization methods and transformation methods. The adversarial examples are crafted by Inc-v3, Inc-v4, IncRes-v2, and Res-101 respectively. * indicates the surrogate model.}
\label{tab:sota}
\footnotesize
\scalebox{0.90}{

\begin{tabularx}{\textwidth}{@{}>{\centering\arraybackslash}X|c|*{10}{>{\centering\arraybackslash}X}@{}}

\toprule
Model & \multicolumn{1}{c|}{Attack} & Inc-v3 & Inc-v4 & Inc-Res-v2 & Res-101 & Inc-v3$_{ens3}$ & Inc-v3$_{ens4}$ & Inc-Res-v2$_{ens}$ & ViT & Swin-b & Avg. \\ 
\midrule
\multirow{13}{*}{\shortstack{Inc\\-v3}} 
 & \multicolumn{1}{c|}{SI-NI-FGSM} & \textbf{100*} & 78.8 & 76.8 & 69.1 & 34.6 & 31.4 & 16.7 & 23.2 & 22.2 & 50.3 \\
\multicolumn{1}{c|}{} & \multicolumn{1}{c|}{RAP-SI-TI-DIM} & \textbf{100*} & 81.0 & 70.6 & 70.2 & 47.2 & 42.8 & 32.6 & 23.2 & 25.0 & 54.7 \\
\multicolumn{1}{c|}{} & \multicolumn{1}{c|}{VT-TI-DIM} & 98.9* & 78.1 & 73.9 & 66.8 & 58.8 & 56.3 & 43.1 & 32.3 & 29.3 & 59.7 \\
\multicolumn{1}{c|}{} & \multicolumn{1}{c|}{SI-TI-DIM} & 98.7* & 85.4 & 80.6 & 76.0 & 64.1 & 62.1 & 45.2 & 35.8 & 29.2 & 64.1 \\
\multicolumn{1}{c|}{} & \multicolumn{1}{c|}{Admix-SI-TI-DIM} & 99.9* & 90.2 & 86.4 & 82.5 & 73.7 & 68.0 & 53.8 & 40.9 & 34.2 & 70.0 \\
\multicolumn{1}{c|}{} & \multicolumn{1}{c|}{BSR-MI-FGSM} & 99.9* & 89.1 & 87.3 & 84.5 & 72.1 & 65.5 & 51.9 & 28.9 & 45.4 & 69.4 \\
\multicolumn{1}{c|}{} & \multicolumn{1}{c|}{SIA-MI-FGSM} & 99.6* & 90.5 & 88.6 & 82.7 & 75.5 & 69.5 & 54.7 & 27.3 & 44.0 & 70.3 \\
\multicolumn{1}{c|}{} & \multicolumn{1}{c|}{SSA-SI-TI-DIM} & 99.8* & 93.5 & 92.3 & 89.4 & 89.7 & 88.1 & 78.5 & 50.5 & 39.4 & 80.1 \\
\multicolumn{1}{c|}{} & \multicolumn{1}{c|}{NGI-SI-TI-DIM(ours)} & 99.4* & 91.7 & 88.8 & 83.6 & 76.8 & 74.1 & 57.7 & 41.6 & 34.9 & 72.1 \\
\multicolumn{1}{c|}{} & \multicolumn{1}{c|}{NGI-Admix-SI-TI-DIM(ours)} & 99.8* & 93.5 & 92.3 & 88.8 & 81.1 & 78.8 & 64.2 & 47.3 & 38.9 & 76.1 \\
\multicolumn{1}{c|}{} & \multicolumn{1}{c|}{NGI-BSR-MI-FGSM(ours)} & \textbf{100*} & 94.5 & 93.7 & 90.2 & 81.5 & 74.4 & 61.0 & 31.3 & \textbf{51.3} & 75.3 \\
\multicolumn{1}{c|}{} & \multicolumn{1}{c|}{NGI-SIA-MI-FGSM(ours)} & \textbf{100*} & 95.5 & 93.2 & 89.9 & 83.4 & 76.3 & 64.3 & 33.1 & 51.0 & 76.3 \\
\multicolumn{1}{c|}{} & \multicolumn{1}{c|}{NGI-SSA-SI-TI-DIM(ours)} & 99.9* & \textbf{96.4} & \textbf{95.3} & \textbf{92.8} & \textbf{91.9} & \textbf{89.9} & \textbf{83.2} & \textbf{56.1} & 42.9 & \textbf{83.2} \\ \midrule
\multirow{13}{*}{\shortstack{Inc\\-v4}} 
 & \multicolumn{1}{c|}{SI-NI-FGSM} & 87.0 & \textbf{100*} & 81.4 & 73.4 & 45.5 & 41.6 & 26.4 & 28.9 & 31.7 & 57.3 \\
\multicolumn{1}{c|}{} & \multicolumn{1}{c|}{RAP-SI-TI-DIM} & 74.7 & \textbf{100*} & 66.6 & 63.0 & 37.2 & 37.6 & 26.7 & 21.7 & 22.7 & 50.0 \\
\multicolumn{1}{c|}{} & \multicolumn{1}{c|}{VT-TI-DIM} & 82.2 & 98.0* & 75.9 & 67.7 & 59.3 & 57.4 & 49.6 & 37.3 & 35.5 & 62.5 \\
\multicolumn{1}{c|}{} & \multicolumn{1}{c|}{SI-TI-DIM} & 87.2 & 98.6* & 83.3 & 78.3 & 72.2 & 67.2 & 57.3 & 39.7 & 36.6 & 68.9 \\
\multicolumn{1}{c|}{} & \multicolumn{1}{c|}{Admix-SI-TI-DIM} & 88.3 & 99.1* & 86.3 & 80.3 & 73.2 & 69.5 & 60.0 & 44.1 & 38.9 & 71.1 \\
\multicolumn{1}{c|}{} & \multicolumn{1}{c|}{BSR-MI-FGSM} &  85.8 & 99.7* & 76.0 & 74.1 & 59.3 & 51.5 & 43.0 & 22.4 & 39.1 & 61.2 \\
\multicolumn{1}{c|}{} & \multicolumn{1}{c|}{SIA-MI-FGSM} & 87.1 & 99.5* & 78.9 & 75.5 & 63.1 & 55.9 & 45.0 & 21.8 & 39.5 & 62.9 \\
\multicolumn{1}{c|}{} & \multicolumn{1}{c|}{SSA-SI-TI-DIM} & 95.4 & 99.2* & 93.4 & 90.0 & 88.8 & 86.5 & 81.5 & 58.6 & 48.6 & 82.4 \\
\multicolumn{1}{c|}{} & \multicolumn{1}{c|}{NGI-SI-TI-DIM(ours)} & 92.1 & 99.1* & 89.4 & 83.9 & 79.3 & 77.2 & 66.6 & 48.8 & 44.0 & 75.6 \\
\multicolumn{1}{c|}{} & \multicolumn{1}{c|}{NGI-Admix-SI-TI-DIM(ours)} & 93.8 & 99.3* & 91.2 & 86.3 & 79.3 & 76.3 & 66.5 & 50.3 & 44.4 & 76.4 \\
\multicolumn{1}{c|}{} & \multicolumn{1}{c|}{NGI-BSR-MI-FGSM(ours)}& 90.7 & 99.8* & 84.1 & 81.3 & 67.4 & 57.4 & 47.7 & 25.5 & 44.2 & 66.5\\
\multicolumn{1}{c|}{} & \multicolumn{1}{c|}{NGI-SIA-MI-FGSM(ours)} & 92.0 & 99.7* & 86.0 & 81.4 & 70.6 & 63.5 & 51.5 & 26.7 & 42.1 & 68.2 \\
\multicolumn{1}{c|}{} & \multicolumn{1}{c|}{NGI-SSA-SI-TI-DIM(ours)} & \textbf{97.2} & 99.5* & \textbf{96.1} & \textbf{93.9} & \textbf{91.2} & \textbf{90.3} & \textbf{84.3} & \textbf{61.9} & \textbf{50.8} & \textbf{85.0} \\ \midrule
\multirow{13}{*}{\shortstack{IncRes\\-v2}} 
& \multicolumn{1}{c|}{SI-NI-FGSM} & 87.8 & 82.2 & 99.4* & 77.6 & 53.2 & 45.4 & 37.1 & 29.1 & 29.1 & 60.1 \\
\multicolumn{1}{c|}{} & \multicolumn{1}{c|}{RAP-SI-TI-DIM} & 77.4 & 77.7 & \textbf{100*} & 64.8 & 45.5 & 39.9 & 37.0 & 22.1 & 21.9 & 54.0 \\
\multicolumn{1}{c|}{} & \multicolumn{1}{c|}{VT-TI-DIM} & 79.4 & 77.4 & 94.5* & 71.2 & 64.9 & 59.9 & 59.0 & 40.8 & 36.0 & 64.8 \\
\multicolumn{1}{c|}{} & \multicolumn{1}{c|}{SI-TI-DIM} & 88.0 & 85.5 & 97.5* & 81.6 & 76.0 & 71.5 & 70.2 & 47.4 & 39.6 & 73.0 \\
\multicolumn{1}{c|}{} & \multicolumn{1}{c|}{Admix-SI-TI-DIM} & 90.3 & 88.7 & 97.7* & 86.1 & 80.6 & 76.2 & 73.2 & 50.8 & 41.5 & 76.1 \\
\multicolumn{1}{c|}{} & \multicolumn{1}{c|}{BSR-MI-FGSM} &  84.9 & 80.3 & 98.4* & 76.1 & 65.6 & 58.0 & 52.8 & 24.0 & 38.7 & 64.3 \\
\multicolumn{1}{c|}{} & \multicolumn{1}{c|}{SIA-MI-FGSM} & 87.1 & 83.5 & 99.1* & 79.7 & 71.7 & 65.4 & 60.0 & 26.6 & 41.8 & 68.3 \\
\multicolumn{1}{c|}{} & \multicolumn{1}{c|}{SSA-SI-TI-DIM} & 93.6 & 93.1 & 97.6* & 91.8 & 91.0 & 90.4 & 89.0 & 61.0 & 49.2 & 84.1 \\
\multicolumn{1}{c|}{} & \multicolumn{1}{c|}{NGI-SI-TI-DIM(ours)} & 92.0 & 91.2 & 97.9* & 89.1 & 84.4 & 82.8 & 81.1 & 57.0 & 48.7 & 80.5 \\
\multicolumn{1}{c|}{} & \multicolumn{1}{c|}{NGI-Admix-SI-TI-DIM(ours)} & 94.1 & 92.9 & 98.8* & 89.7 & 86.4 & 84.5 & 82.0 & 58.0 & 48.2 & 81.6 \\
\multicolumn{1}{c|}{} & \multicolumn{1}{c|}{NGI-BSR-MI-FGSM(ours)}& 88.9 & 85.4 & 98.9* & 81.7 & 72.4 & 64.2 & 59.8 & 28.9 & 44.4 & 69.4\\
\multicolumn{1}{c|}{} & \multicolumn{1}{c|}{NGI-SIA-MI-FGSM(ours)} & 92.5 & 89.2 & 99.5* & 85.3 & 77.6 & 70.1 & 68.0 & 29.3 & 45.3 & 73.0 \\
\multicolumn{1}{c|}{} & \multicolumn{1}{c|}{NGI-SSA-SI-TI-DIM(ours)} & \textbf{96.6} & \textbf{95.8} & 98.6* & \textbf{94.2} & \textbf{94.1} & \textbf{93.4} & \textbf{92.6} & \textbf{65.4} & \textbf{53.2} & \textbf{87.1} \\ \midrule
\multirow{13}{*}{\shortstack{Res\\-101}} 
& \multicolumn{1}{c|}{SI-NI-FGSM} & 82.0 & 77.2 & 76.2 & 99.8* & 43.8 & 39.0 & 25.1 & 26.2 & 24.3 & 54.8 \\
\multicolumn{1}{c|}{} & \multicolumn{1}{c|}{RAP-SI-TI-DIM} & 82.8 & 83.2 & 65.2 & \textbf{100*} & 56.1 & 50.0 & 43.8 & 31.3 & 31.4 & 60.4 \\
\multicolumn{1}{c|}{} & \multicolumn{1}{c|}{VT-TI-DIM} & 81.3 & 78.2 & 77.6 & 98.5* & 68.3 & 64.5 & 57.4 & 42.7 & 36.3 & 67.2 \\
\multicolumn{1}{c|}{} & \multicolumn{1}{c|}{SI-TI-DIM} & 86.5 & 81.8 & 83.2 & 98.9* & 77.0 & 72.3 & 61.9 & 41.9 & 34.8 & 70.9 \\
\multicolumn{1}{c|}{} & \multicolumn{1}{c|}{Admix-SI-TI-DIM} & 88.6 & 85.0 & 86.0 & 99.8* & 76.9 & 71.9 & 63.3 & 46.4 & 35.3 & 72.6 \\
\multicolumn{1}{c|}{} & \multicolumn{1}{c|}{BSR-MI-FGSM} &  89.0 & 90.3 & 83.1 & 92.9* & 70.0 & 61.6 & 55.2 & 36.6 & 56.7 & 70.6 \\
\multicolumn{1}{c|}{} & \multicolumn{1}{c|}{SIA-MI-FGSM} & 92.1 & 90.5 & 85.6 & 94.3* & 75.5 & 70.7 & 60.9 & 38.8 & 58.6 & 74.1 \\
\multicolumn{1}{c|}{} & \multicolumn{1}{c|}{SSA-SI-TI-DIM} & 91.8 & 92.1 & 89.8 & 94.9* & 86.5 & 85.8 & 79.0 & 57.9 & 44.2 & 80.2 \\
\multicolumn{1}{c|}{} & \multicolumn{1}{c|}{NGI-SI-TI-DIM(ours)} & 90.8 & 86.1 & 87.9 & 99.3* & 82.9 & 79.3 & 72.5 & 50.5 & 39.5 & 76.5 \\
\multicolumn{1}{c|}{} & \multicolumn{1}{c|}{NGI-Admix-SI-TI-DIM(ours)} & 90.9 & 87.3 & 89.1 & 99.8* & 83.9 & 80.5 & 72.4 & 52.3 & 41.6 & 77.5 \\
\multicolumn{1}{c|}{} & \multicolumn{1}{c|}{NGI-BSR-MI-FGSM(ours)}& 94.6 & 94.6 & 89.2 & 96.8* & 78.9 & 70.3 & 63.8 & 41.1 & 62.9 & 76.9\\
\multicolumn{1}{c|}{} & \multicolumn{1}{c|}{NGI-SIA-MI-FGSM(ours)} & \textbf{96.0} & 95.1 & 91.7 & 97.4* & 84.2 & 77.0 & 69.5 & 42.5 & 66.5 & 80.0 \\
\multicolumn{1}{c|}{} & NGI-SSA-SI-TI-DIM(ours) & 95.4 & \textbf{95.6} & \textbf{93.4} & 97.7* & \textbf{89.9} & \textbf{88.7} & \textbf{81.9} & \textbf{60.9} & \textbf{45.5} & \textbf{83.2} \\ \bottomrule
\end{tabularx}

}

\end{table*}

\begin{figure*}[t]
\centering
\includegraphics[width=1\textwidth]{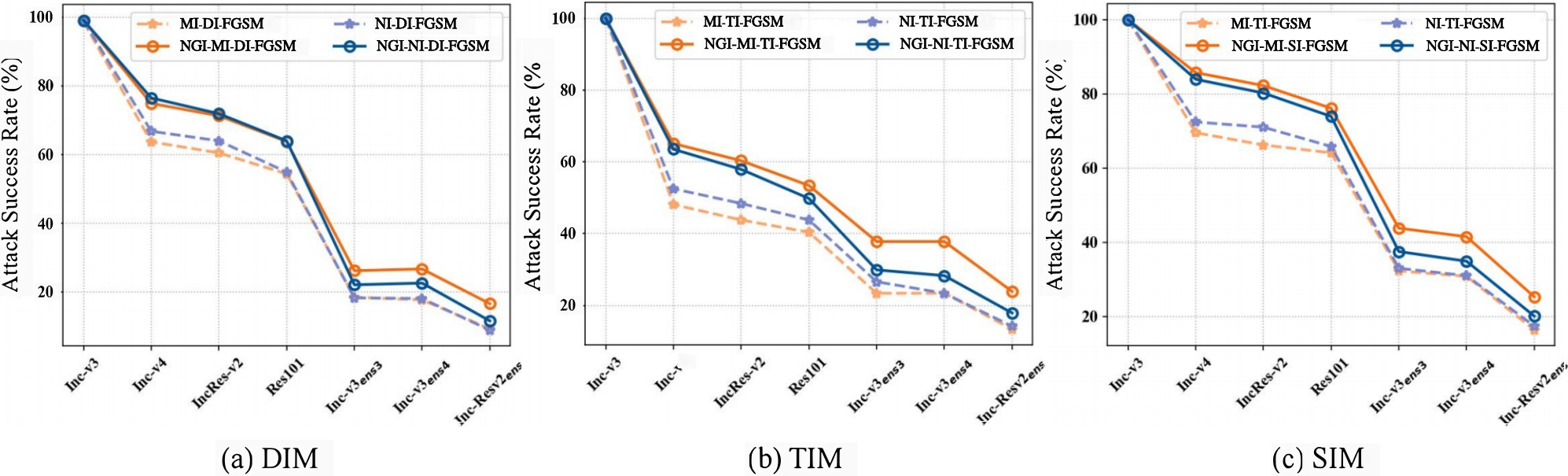}
\caption{Attack success rate (\%) of input transformation-based attacks on seven models in the single-model setting. The adversarial examples are crafted on Inc-v3.}
\label{fig:exp4.1}
\end{figure*}

\begin{table*}[t]
\caption{Attack success rates (\%) of nine normally and adversarially trained models. The adversarial examples are crafted by the ensemble of Inc-v3, Inc-v4, IncRes-v2, and Res-101 models. * indicates the surrogate model.}
\label{tab:ens}
\centering
\begin{tabularx}{\textwidth}{@{}c|*{10}{>{\centering\arraybackslash}X}@{}}
\toprule
Attack & Inc-v3 & Inc-v4 & Inc-Res-v2 & Res-101 & Inc-v3$_{ens3}$ & Inc-v3$_{ens4}$ & IncRes-v2$_{ens}$ & ViT & Swin-b & Avg. \\ \midrule
SI-NI-FGSM & 99.8* & 99.2* & 97.2* & 99.7* & 41.0 & 36.1 & 24.4 & 32.1 & 34.0 & 62.6 \\
VT-TI-DIM & 98.2* & 96.3* & 93.7* & 98.4* & 80.7 & 76.0 & 74.4 & 55.2 & 54.0 & 80.8 \\
SI-TI-DIM & 99.0* & 98.4* & 97.3* & 99.3* & 91.1 & 89.8 & 84.5 & 71.5 & 64.5 & 88.4 \\
Admix-SI-TI-DIM & 99.4* & 98.7* & 97.7* & 99.8* & 91.4 & 88.8 & 85.8 & 70.0 & 65.6 & 88.6 \\
BSR-MI-FGSM & 99.8* & 99.6* & 99.4* & 98.2* & 94.7 & 92.1 & 89.2 & 63.6 & 75.9 & 90.3 \\
SSA-SI-TI-DIM & 99.5* & 98.7* & 98.5* & 97.1* & 96.6 & 96.4 & 95.2 & 80.3 & 69.9 & 92.5 \\
SIA-MI-FGSM & 99.9* & 99.6* & 99.3* & 98.5* & 95.3 & 93.9 & 90.5 & 63.6 & 77.9 & 90.9 \\
NGI-SI-TI-DIM(ours) & 98.8* & 98.5* & 98.0* & 99.0* & 96.1 & 94.8 & 92.8 & 82.8 & 74.2 & 92.8 \\
NGI-Admix-SI-TI-DIM(ours) & 99.7* & 99.4* & 98.8* & \textbf{99.9*} & 96.3 & 95.2 & 93.5 & 82.1 & 75.3 & 93.4 \\
NGI-BSR-MI-FGSM(ours) & \textbf{100*} & 99.8* & \textbf{99.8*} & 99.4* & 97.9 & 97.0 & 93.4 & 70.3 & 82.3 & 93.3 \\
NGI-SSA-SI-TI-DIM(ours) & 99.7* & 99.7* & 99.4* & 98.6* & 98.4 & 97.7 & \textbf{96.9} & \textbf{84.3} & 74.9 & \textbf{94.4} \\
NGI-SIA-MI-FGSM(ours) & \textbf{100*} & \textbf{99.9*} & \textbf{99.8*} & 99.4* & \textbf{98.5} & \textbf{98.0} & 95.8 & 72.2 & \textbf{84.7} & 94.3 \\ \bottomrule
\end{tabularx}
\end{table*}

\textbf{Attack on input transformation-based attacks.}
Input transformation-based methods like DIM, SIM, and TIM, when combined with optimization-based methods, can enhance adversarial example transferability through multiple input transformations. We integrate our strategy with such methods to assess further transferability improvements, as shown in Figure \ref{fig:exp4.1}.

The results reveal that our method effectively improves the 
ASR
for both black-box and white-box models, particularly for black-box models, with an average improvement of 8\%, 13.2\%, and 10.7\% compared to the DIM, TIM, and SIM baselines, respectively. Nevertheless, Lin \textit{et al.}~\citep{DBLP:journals/corr/abs-1908-06281} discover that merging DIM, TIM, and SIM can further enhance the efficiency of optimization-based attacks. We regard this as one of the current state-of-the-art methods and will delve into it in Table \ref{tab:sota}.

\textbf{Compare with state-of-the-art methods}
In this part, we compare our approach with several prominent advanced adversarial attacks that have been recently introduced. By integrating our proposed strategy with five top-performing methods, we examine their performance enhancement in black-box model attacks. As shown in Table \ref{tab:sota}, our approach can significantly enhance the transferability of the adversarial examples, outperforming the competing methods. 
For instance, when using Inc-Resv2 as the surrogate model for generating adversarial examples, our strategy achieves an average of 87.1\%, 73.0\%, 69.4\%, 81.6\%, and 80.5\%, representing an average improvement of 3.0\%, 4.7\%, 5.1\%, 5.5\%, and 7.5\% over SSA-SI-TI-DIM, SIA-MI-FGSM, BSR-MI-FGSM, Admix-SI-TI-DIM, and SI-TI-DIM, respectively. It's worth noting that the recent RAP attack strategy, which claims to achieve superior black-box transferability with more steps, only reached an average ASR of 54.8\% for RAP-SI-TI-DIM. Conversely, NGI-SI-TI-DIM attained a significantly higher average ASR of 76.2\%, which substantiates the effectiveness of our approach.

\subsubsection{Evaluation on Ensemble-model Attacks} 
The strategy of employing ensemble models has been validated to significantly enhance the transferability of adversarial examples~\citep{DBLP:conf/iclr/LiuCLS17}. In this section, we elucidate the effectiveness of our method within this attack paradigm. Specifically, we generate adversarial examples by attacking multiple surrogate models, \textit{e.g.}, Inc-v3, Inc-v4, Inc-Resv2, and Res-101, and subsequently evaluate their performance within chosen target models. 
Table~\ref{tab:ens} demonstrates that our strategy achieves an average ASR of 92.8\%, 93.3\%, 93.4\%, 94.4\%, and 94.3\% under the ensemble-model setting. 
Further contrasting this with established methods like SI-TI-DIM, BSR-MI-FGSM, Admix-SI-TI-DIM, SSA-SI-TI-DIM, and SIA-MI-FGSM our approach exhibits a marked enhancement in transferability. Specifically, for adversarially trained models, the ASR ranges from 92.8\% to 98.5\%. Additionally, for models employing the transformer architecture, there is a remarkable increase in transferability, ranging from 4.0\% to 12.1\%.

\begin{table*}[]
\centering
\caption{Attack success rates (\%) of nine advance adversarial defenses. The adversarial examples are crafted by the ensemble of Inc-v3, Inc-v4, IncRes-v2, and Res-101 models.}
\label{tab:defense}

\begin{tabularx}{\textwidth}{@{}c|*{11}{>{\centering\arraybackslash}X}@{}}
\toprule
Attack & HGD & R\&P & NIPS-r3 & JPEG & FD & Com-Defend & RS & Bit-Red & NRP & DiffPure & Avg. \\ \midrule
SI-TI-DIM & 90.4 & 86.3 & 86.7 & 91.5 & 89.1 & 90.0 & 68.6 & 76.0 & 75.1 & 61.6 & 81.5 \\
NGI-SI-TI-DIM(ours) & \textbf{94.7} & \textbf{93.8} & \textbf{93.7} & \textbf{96.3} & \textbf{93.8} & \textbf{94.9} & \textbf{78.5} & \textbf{84.2} & \textbf{86.9} & \textbf{69.8} & \textbf{88.7} \\ \midrule
Admix-SI-TI-DIM & 90.8 & 87.1 & 87.3 & 92.0 & 88.3 & 89.4 & 68.1 & 73.7 & 73.8 & 61.0 & 81.2 \\
NGI-Admix-SI-TI-DIM(ours) & \textbf{95.3} & \textbf{93.9} & \textbf{93.5} & \textbf{96.5} & \textbf{92.7} & \textbf{93.9} & \textbf{78.5} & \textbf{84.9} & \textbf{86.5} & \textbf{71.7} & \textbf{88.7} \\ \midrule
SSA-SI-TI-DIM & 95.9 & 95.4 & 95.3 & 96.8 & 95.1 & 96.2 & 90.3 & 89.5 & 90.7 & 84.3 & 93.0 \\
NGI-SSA-SI-TI-DIM(ours) & \textbf{97.5} & \textbf{97.1} & \textbf{97.4} & \textbf{98.4} & \textbf{97.0} & \textbf{97.5} & \textbf{93.2} & \textbf{91.9} & \textbf{92.2} & \textbf{89.6} & \textbf{95.2} \\ \bottomrule
\end{tabularx}
\end{table*}

\subsubsection{Evaluation on Defense Models}
To thoroughly assess our approach, we evaluate its performance against ten recent adversarial defense strategies, including the top three from the NIPS defense competition (HGD~\citep{DBLP:conf/cvpr/LiaoLDPH018}, R\&P~\citep{xie2017mitigating}, NIPS-r3\footnote{\url{https://github.com/anlthms/nips-2017/tree/master/mmd}}), along with seven advanced methods (Bit-Red~\citep{DBLP:conf/ndss/Xu0Q18}, JPEG~\citep{guo2017countering}, FD~\citep{liu2019feature}, ComDefend~\citep{DBLP:conf/cvpr/JiaWCF19}, RS~\citep{DBLP:conf/icml/CohenRK19}, NRP~\citep{DBLP:conf/cvpr/NaseerKHKP20}), DiffPure~\citep{nie2022DiffPure}.
As depicted in Table \ref{tab:defense}, our approach exhibits strong attack performance against multiple defense strategies, notably the RS defense. While SI-TI-DIM and Admix-SI-TI-DIM achieve ASR of 68.6\% and 68.1\% against the RS defense, respectively, our strategy significantly enhances the transferability of adversarial examples, yielding improvements of 9.9\% and 10.4\% over these baselines. Furthermore, against the robust defense mechanism recently introduced as Neural Representation Purifier (NRP), our approach successfully attains an impressive ASR exceeding 86.5\%. 
We further assess our approach under the recent state-of-the-art adversarial purification method, DiffPure. Despite DiffPure's robust defense, our strategy achieves notable attack success rates of 69.8\%, 71.7\%, and 89.6\%.
Moreover, our method consistently maintains remarkable attack success rates, averaging 88.7\%, 88.7\%, and 95.2\%, further demonstrating its effectiveness.

\begin{table*}[t]
\centering
\caption{Attack success rates (\%) of baseline combining different proposed modules on seven models in the single-model setting. The adversarial examples are crafted on Inc-v3. * indicates the surrogate model. "w/" denotes "with".}
\label{tab:ablation1}

\begin{tabular}{@{}c|cccccccc@{}}
\toprule
Attack & Inc-v3 & Inc-v4 & Inc-Res-v2 & Res-101 & Inc-v3$_{ens3}$ & Inc-v3$_{ens4}$ & IncRes-v2$_{ens}$ & Avg. \\ 
\midrule
Baseline & 99.4* & 83.6 & 80.6 & 74.8 & 64.4 & 62.7 & 43.9 & 72.8 \\
w/ EB & 99.1* & 86.2 & 83.7 & 80.3 & 66.0 & 62.5 & 45.2 & 74.7 \\
w/ MM & 98.5* & 81.6 & 79.0 & 75.6 & 67.0 & 66.3 & 51.1 & 74.2 \\
w/ EB, MM & \textbf{99.4*} & \textbf{91.7} & \textbf{88.8} & \textbf{83.6} & \textbf{76.8} & \textbf{74.1} & \textbf{57.7} & \textbf{81.7} \\ 
\bottomrule
\end{tabular}
\end{table*}

\subsection{Ablation Study}
We conduct a series of ablation studies on the proposed NGI-Attack, aiming to investigate the effects of different proposed modules and hyperparameters. 
In these experiments, we combine the proposed strategy with SI-TI-DIM, \textit{i.e.}, NGI-SI-TI-DIM, and then employ it on Inc-v3 to generate adversarial examples, which are later tested on the target models.

{
\subsubsection{Effect of different modules}
In this section, we analyze the proposed modules, namely Example Backtracking (EB) and Multiplex Mask (MM), and their contributions to improving the transferability of adversarial examples. The average results are presented in Table~\ref{tab:ablation1}.
We apply SI-TI-DIM as our baseline.
We first investigate Example Backtracking and directly increase the step size in the second stage to generate sufficient perturbation. The results, \textit{i.e.} w/ EB, demonstrate that Example Backtracking generally enhances the transferability of adversarial examples, regardless of whether the models are trained normally or adversarially. 
On this basis, we introduce the Multiplex Mask module, aiming to provide richer gradient information to compensate for the limitation of directly increasing step size. 
To delve deeper into the relationship between the mask branch and the transferability of adversarial examples, we exclusively adopt attacks utilizing the mask branch, as demonstrated with the 'w/ MM'. 
It is evident that the adversarial examples generated from images processed by MaskProcess significantly improve the attack success rate on adversarially trained models. We attribute this to the ability of MaskProcess to guide the network to pay more attention to non-discriminative regions, thereby launching a more comprehensive attack.
However, images processed by MaskProcess indeed lose a portion of the original information. We hypothesize that this loss is the underlying cause for the performance degradation in white-box attacks and when transferring to normally trained models.
Overall, when incorporating EB and MM, we can effectively raise the transferability of adversarial examples to 81.7\% 
}

\begin{table*}[t]
\centering
\caption{Attack success rates (\%) of baseline combining different proposed modules on seven models in the single-model setting. The adversarial examples are crafted on Inc-v3. * indicates the surrogate model. }
\label{tab:abl_trans}
\small
\begin{tabular}{@{}ccccccccc@{}}
\toprule
Attack & Inc-v3 & Inc-v4 & Inc-Res-v2 & Res-101 & Inc-v3$_{ens3}$ & Inc-v3$_{ens4}$ & IncRes-v2$_{ens}$ & Avg. \\ \midrule
EB+multi-way (Clean) & 98.9* & 90.8 & 86.8 & 84.2 & 66.1 & 61.3 & 44.1 & 76.0 \\
EB+multi-way (DI) & 99.2* & \textbf{91.8} & 88.3 & \textbf{84.4} & 74.6 & 71.6 & 53.9 & 80.5 \\
EB+multi-way (MaskProcess) & \textbf{99.4*} & 91.7 & \textbf{88.8} & 83.6 & \textbf{76.8} & \textbf{74.1} & \textbf{57.7} & \textbf{81.7} \\ \bottomrule
\end{tabular}
\end{table*}

\begin{figure*}[t]
\centering
\subfloat[Ablation study on hyperparameter of Example Backtracking Steps $K$.]{
  \includegraphics[width=0.33\textwidth]{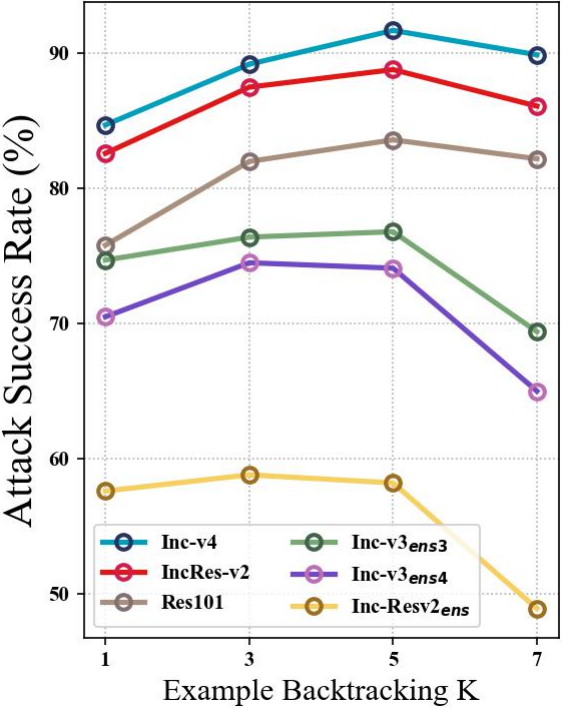}
  \label{fig:abl1}
}
\hspace{1em}
\subfloat[Ablation study on hyperparameter of Multiplex Mask Probability $P$.]{
  \includegraphics[width=0.59\textwidth]{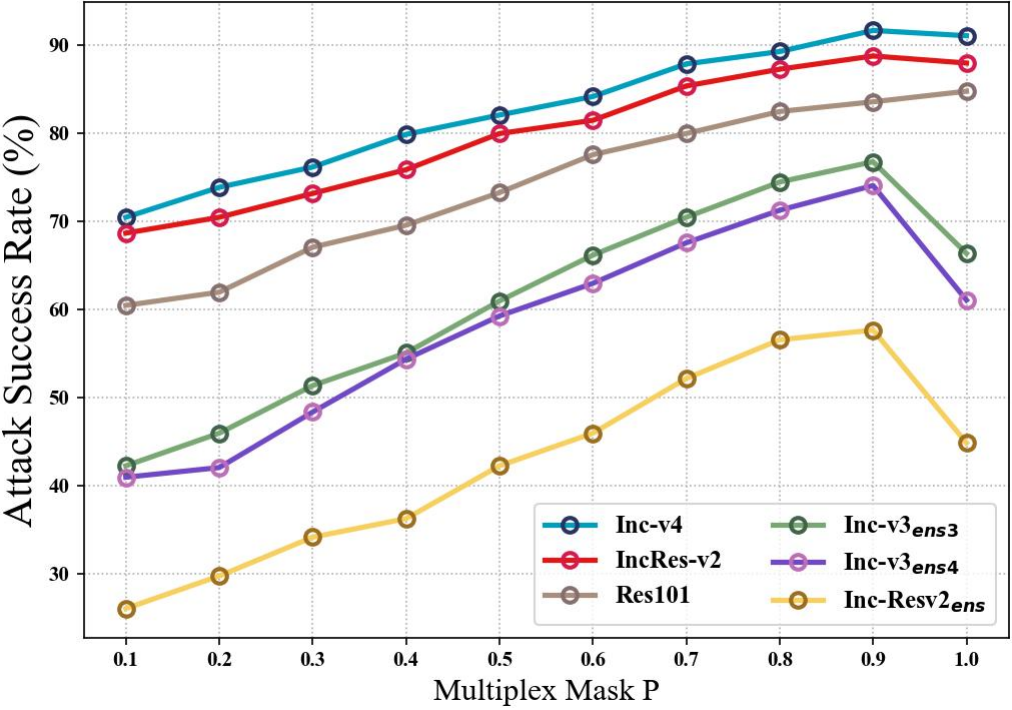}
  \label{fig:abl2}
}
\caption{Ablation analysis on key hyperparameters in our proposed modules.}
\label{fig:ablation}
\end{figure*}

\subsubsection{Effect of different input transformation strategies}
In this section, we evaluate the effectiveness of different input transformation strategies within our multi-way attack strategy.
To fully leverage the Neighborhood Gradient Information (NGI) without incurring additional time costs, our strategy involves increasing attack step in subsequent iterations to achieve sufficient perturbation. However, we observed that directly increasing the attack step often yields poor results on adversarially trained models, which typically have a more extensive attention region ~\citep{dong2019evading}. Additionally, our studies found that the MaskProcess strategy encourages models to focus more on non-discriminative regions.
In the MM process, \textit{i.e.}, the multi-way attack phase, acquiring diverse gradient information effectively prevents overfitting that could result from directly increasing the step size. To validate whether MaskProcess focuses on non-discriminative regions, we conduct ablation experiments comparing the effects of different input transformation methods in multi-way attacks. Specifically, we compare the use of clean inputs, perturbed inputs (DI inputs), and MaskProcess inputs in one branch. The results are shown in Table \ref{tab:abl_trans}. The results show that under normally trained models, the transferability of adversarial examples generated from MaskProcess inputs is comparable to that from DI inputs; however, under adversarially trained models, MaskProcess inputs significantly outperform DI inputs, showing an improvement of 2.2\% to 3.8\% in attack success rate.
In conclusion, our multi-way attack strategy can significantly enhance the adversarial transferability across various models, particularly demonstrating advantages in adversarially trained models.

\subsubsection{Example Backtracking Step \textit{$K$}}
We first explore the impact of the hyperparameter $K$ during the Example Backtracking stage on the transferability of adversarial examples. In this phase, we collect information on adjacent gradients for a range of different $K$ values. As demonstrated in Figure~\ref{fig:abl1}, we select $K$ values of 1, 3, 5, and 7 for analysis. The results indicate that when the $K$ value is small ($K=1$), the transferability is not strong enough, which we suspect is due to the insufficient gathering of adjacent gradient information during this stage. However, as the $K$ value increases (\textit{e.g.}, $K=3$ or $K=5$), the success rate of the attack significantly improves. On the contrary, we discover that as $K$ continues to escalate, the transferability of adversarial examples cannot be further enhanced. We speculate this is due to adversarial examples deviating significantly from clean images, thereby failing to provide useful gradient information near the clean images. In conclusion, we choose $K=5$ in our experiments.

\subsubsection{Multiplex Mask Probability \textit{$P$}}
We subsequently examine the effects of pixel masking at varying probabilities on auxiliary pathways in the Multiplex Mask stage.  As depicted in Figure~\ref{fig:abl2}, we conduct a series of experiments with incremental $P$ values ranging from 0.1 to 1.0. The findings indicate that adversarial example efficacy is diminished when $P$ is small, signifying that each pixel in the image is retained at a relatively low probability. We infer that the poor performance results from the substantial input disparity between the two pathways. As the $P$ value increases, the attack success rate on the target models notably amplifies, peaking particularly when $P = 0.9$. We attribute this enhancement to the auxiliary pathway's capability to provide more comprehensive gradient information based on the randomly masked images. However, when $P$ reaches 1.0, the Multiplex Mask stage degrades into a direct step size change, which could potentially lead to overfitting during the attack process.
\color{black}

\begin{table*}[t]
\centering
\caption{Comparison of average ASR with standard deviation (\%) on nine target models. Each result is averaged over five random trials using Inception-v3 as the surrogate model.}
\label{tab:std_results}
\small
\scalebox{0.7}{
\begin{tabular}{@{}c|ccccccccc@{}}
\toprule
Attack & Inc-v3 & Inc-v4 & Inc-Res-v2 & Res-101 & Inc-v3$_{\text{ens3}}$ & Inc-v3$_{\text{ens4}}$ & IncRes-v2$_{\text{ens}}$ & ViT & Swin-B \\
\midrule
MI-FGSM & \textbf{100.00\,($\pm$\,0.00)} & 44.34\,($\pm$\,0.46) & 41.56\,($\pm$\,0.42) & 36.16\,($\pm$\,0.33) & 14.00\,($\pm$\,0.33) & 12.82\,($\pm$\,0.58) & 6.22\,($\pm$\,0.31) & 7.88\,($\pm$\,0.34) & 10.12\,($\pm$\,0.53) \\ \midrule
NGI-MI-FGSM (ours) & \textbf{100.00\,($\pm$\,0.00)} & \textbf{62.82\,($\pm$\,0.75)} & \textbf{59.34\,($\pm$\,0.41)} & \textbf{50.16\,($\pm$\,0.93)} & \textbf{21.06\,($\pm$\,0.48)} & \textbf{20.62\,($\pm$\,0.60)} & \textbf{11.16\,($\pm$\,0.42)} & \textbf{11.54\,($\pm$\,0.51)} & \textbf{15.58\,($\pm$\,0.31)} \\ \bottomrule
\end{tabular}}
\end{table*}
\renewcommand{\thetable}{\arabic{table}}

\begin{table*}[h]
\centering
\caption{Bootstrap hypothesis test (10,000 resamples, 5 independent runs) comparing the ASR of \textbf{NGI-MI-FGSM} with its baseline \textbf{MI-FGSM} on eight black-box target models.  One-sided $p$-values are reported; a $p$-value below 0.05 indicates that the improvement of NGI-MI-FGSM is statistically significant.}
\label{tab:significance}
\normalsize
\scalebox{0.9}{
\begin{tabular}{@{}ccc@{}}
\toprule
Model & Bootstrap p-value & Significant (p \textless 0.05) \\ \midrule
Inc-v4 & 0.0057 & TRUE \\
Inc-Res-v2 & 0.0038 & TRUE \\
Res-101 & 0.005 & TRUE \\
Inc-v3$_{ens3}$ & 0.001 & TRUE \\
Inc-v3$_{ens4}$ & 0.002 & TRUE \\
IncRes-v2$_{ens}$ & 0.0018 & TRUE \\
Vit & 0.0001 & TRUE \\
Swin-b & 0.0015 & TRUE \\ \bottomrule
\end{tabular}}
\end{table*}

\subsection{Statistical Robustness and Significance Validation}
To enhance the credibility and statistical robustness of our evaluation, we perform five independent runs for each experimental setting using different random seeds. 
As shown in Table~\ref{tab:std_results}, we report the average ASR along with the standard deviation on eight black-box models when using Inception-v3 as the surrogate model.

We observe that attacks incorporating NGI-Attack exhibit slightly higher standard deviations compared to the baseline methods. 
This is primarily due to the stochastic nature of the Multiplex Mask strategy adopted in the second stage of our NGI-Attack. 
The random masking of inputs introduces greater diversity in gradient information, which improves transferability but also increases variance across trials. 
Nonetheless, NGI consistently outperforms the baseline in terms of average ASR, validating its effectiveness.

To further validate the statistical significance of our improvements, we conduct a bootstrap hypothesis test~\citep{bootstrap} with 10,000 resampling iterations. 
Specifically, we compare the ASR results of Table~\ref{tab:std_results} and calculate one-sided p-values for each black-box target model.
As presented in Table~\ref{tab:significance}, all p-values are below the 0.05 threshold, confirming that the improvement of NGI-MI-FGSM over MI-FGSM is statistically significant. 
This analysis provides strong empirical evidence supporting the robustness and reliability of our method.

\section{Conclusions}
\label{sec:conclusion}
In this work, we validate that the Neighbourhood Gradient Information exhibits stronger transferability. To fully use this gradient information, we introduce a novel 
NGI-Attack, incorporating two key strategies: Example Backtracking and Multiplex Mask. 
Specifically, Example Backtracking effectively accumulates gradient information around clean images, which can be utilized efficiently in the second phase of the attack. 
Then, we employ Multiplex Mask strategy, which can effectively scatter the network's attention and force network focus on non-discriminative regions. This strategy not only provides richer gradient information but also achieves sufficient perturbation without incurring additional time costs.
Extensive experiments demonstrate that our proposed strategy is effective in enhancing the transferability of existing attack methods, even in the face of various advanced defense methods.

{\textbf{Limitation.} While NGI-Attack significantly enhances adversarial transferability, it still exhibits certain limitations. First, its performance may degrade when there is a substantial architectural gap between the surrogate and target models, as the NGI may fail to align well with the target model’s decision boundaries. Second, a significant domain gap between source and target data can also reduce the effectiveness of the attack. 
These limitations suggest promising directions for improving the robustness and generalizability of NGI-based attacks.}

\textbf{Broader Impact.}
In this study, we present a novel NGI-Attack, incorporating
Example Backtracking and Multiplex Mask strategies. These methods exhibit substantial transferability potential under black-box attack scenarios, while concurrently emphasizing the inherent vulnerabilities of DNNs. This revelation highlights the necessity for intensive research aimed at enhancing network robustness, offering significant insights into building secure and dependable neural networks.

\textbf{Future Work.} Moreover, since NGI-Attack is fundamentally a gradient-based strategy, its core principles are not restricted to vision models. Future work could explore its extension to other modalities such as natural language processing and speech, where gradients also play a central role in model behavior.

\section*{CRediT authorship contribution statement}
Haijing Guo: Conceptualization, Investigation, Experimentation, Visualization, Writing-Original Draft, Review \& Editing.
Jiafeng Wang: Conceptualization, Investigation, Experimentation, Visualization, Writing-Original Draft, Review \& Editing.
Zhaoyu Chen: Investigation, Experimentation, Review \& Editing, Supervision.
Kaixun Jiang: Investigation, Experimentation, Review \& Editing.
Lingyi Hong: Investigation, Review \& Editing.
Pinxue Guo: Review \& Editing.
JingLun Li: Review \& Editing.
Wenqiang Zhang: Supervision, Review \& Editing, Funding acquisition.

\section*{Declaration of competing interest}
The authors declare that they have no known competing financial
interests or personal relationships that could have appeared to influence
the work reported in this paper. 

\section*{Data availability}
Data will be made available on request. 

\section*{Acknowledgments}
This work was supported by National Natural Science Foundation of China (No.62072112), Scientific and Technological innovation action plan of Shanghai Science and Technology Committee (No.22511102202), Fudan Double First-class Construction Fund (No. XM03211178).

\bibliographystyle{unsrt}
\bibliography{cas-refs}
\end{document}